# EVALUATION: FROM PRECISION, RECALL AND F-MEASURE TO ROC, INFORMEDNESS, MARKEDNESS & CORRELATION

**POWERS, D.M.W.**
*AILab, School of Computer Science, Engineering and Mathematics, Flinders University, South Australia, Australia
Corresponding author. Email: David.Powers@flinders.edu.au

**Abstract**-Commonly used evaluation measures including Recall, Precision, F-Measure and Rand Accuracy are biased and should not be used without clear understanding of the biases, and corresponding identification of chance or base case levels of the statistic. Using these measures a system that performs worse in the objective sense of Informedness, can appear to perform better under any of these commonly used measures. We discuss several concepts and measures that reflect the probability that prediction is informed versus chance. Informedness and introduce Markedness as a dual measure for the probability that prediction is marked versus chance. Finally we demonstrate elegant connections between the concepts of Informedness, Markedness, Correlation and Significance as well as their intuitive relationships with Recall and Precision, and outline the extension from the dichotomous case to the general multi-class case.

## INTRODUCTION

A common but poorly motivated way of evaluating results of Machine Learning experiments is using Recall, Precision and F-measure. These measures are named for their origin in Information Retrieval and present specific biases, namely that they ignore performance in correctly handling negative examples, they propagate the underlying marginal prevalences and biases, and they fail to take account the chance level performance. In the Medical Sciences, Receiver Operating Characteristics (ROC) analysis has been borrowed from Signal Processing to become a standard for evaluation and standard setting, comparing True Positive Rate and False Positive Rate. In the Behavioural Sciences, Specificity and Sensitivity, are commonly used. Alternate techniques, such as Rand Accuracy and Cohen Kappa, have some advantages but are nonetheless still biased measures. We will recapitulate some of the literature relating to the problems with these measures, as well as considering a number of other techniques that have been introduced and argued within each of these fields, aiming/claiming to address the problems with these simplistic measures.

This paper recapitulates and re-examines the relationships between these various measures, develops new insights into the problem of measuring the effectiveness of an empirical decision system or a scientific experiment, analyzing and introducing new probabilistic and information theoretic measures that overcome the problems with Recall, Precision and their derivatives.

## THE BINARY CASE

It is common to introduce the various measures in the context of a dichotomous binary classification problem, where the labels are by convention + and - and the predictions of a classifier are summarized in a four-cell contingency table. This contingency table may be expressed using raw counts of the number of times each predicted label is associated with each real class, or may be expressed in relative terms. Cell and margin labels may be formal probability expressions, may derive cell expressions from margin labels or vice-versa, may use alphabetic constant labels `a`, `b`, `c`, `d` or `A`, `B`, `C`, `D`, or may use acronyms for the generic terms for True and False, Real and Predicted Positives and Negatives. Often UPPER CASE is used where the values are counts, and lower case letters where the values are probabilities or proportions relative to `N` or the marginal probabilities – we will adopt this convention throughout this paper (always written in `typewriter font`), and in addition will use Mixed Case (in the normal text font) for popular nomenclature that may or may not correspond directly to one of our formal systematic names. True and False Positives (`TP/FP`) refer to the number of Predicted Positives that were correct/incorrect, and similarly for True and False Negatives (`TN/FN`), and these four cells sum to `N`. On the other hand `tp`, `fp`, `fn`, `tn` and `rp`, `rn` and `pp`, `pn` refer to the joint and marginal probabilities, and the four contingency cells and the two pairs of marginal probabilities each sum to 1. We will attach other popular names to some of these probabilities in due course.

We thus make the specific assumptions that we are predicting and assessing a single condition that is either positive or negative (dichotomous), that we have one predicting model, and one gold standard labeling. Unless otherwise noted we will also for simplicity assume that the contingency is non-trivial in the sense that both positive and negative states of both predicted and real conditions occur, so that none of the marginal sums or probabilities is zero.

We illustrate in Table 1 the general form of a binary contingency table using both the traditional alphabetic notation and the directly interpretable systematic approach. Both definitions and derivations in this paper are made relative to these labellings, although English terms (e.g. from Information Retrieval) will also be introduced for various ratios and probabilities. The green positive diagonal represents correct predictions, and the pink negative diagonal incorrect predictions. The predictions of the contingency table may be the predictions of a theory, of some computational rule or system (e.g. an Expert System or a Neural Network), or may simply be a direct measurement, a calculated metric, or a latent condition, symptom or marker. We will refer generically to "the model" as the source of the predicted labels, and "the population" or "the world" as the source of the real conditions. We are interested in understanding to what extent the model "informs" predictions about the world/population, and the world/population "marks" conditions in the model.

**Recall & Precision, Sensitivity & Specificity**

Recall or Sensitivity (as it is called in Psychology) is the proportion of Real Positive cases that are correctly Predicted Positive. This measures the Coverage of the Real Positive cases by the **+P** (Predicted Positive) rule. Its desirable feature is that it reflects how many of the relevant cases the **+P** rule picks up. It tends not to be very highly valued in Information Retrieval (on the assumptions that there are many relevant documents, that it doesn't really matter which subset we find, that we can't know anything about the relevance of documents that aren't returned). Recall tends to be neglected or averaged away in Machine Learning and Computational Linguistics (where the focus is on how confident we can be in the rule or classifier). However, in a Computational Linguistics/Machine Translation context Recall has been shown to have a major weight in predicting the success of Word Alignment [1]. In a Medical context Recall is moreover regarded as primary, as the aim is to identify all Real Positive cases, and it is also one of the legs on which ROC analysis stands. In this context it is referred to as True Positive Rate (`tpr`). Recall is defined, with its various common appellations, by equation (1):

Recall = Sensitivity = `tpr` = tp/rp
= TP / RP = A / (A+C)     (1)

Conversely, Precision or Confidence (as it is called in Data Mining) denotes the proportion of Predicted Positive cases that are correctly Real Positives. This is what Machine Learning, Data Mining and Information Retrieval focus on, but it is totally ignored in ROC analysis. It can however analogously be called True Positive Accuracy (`tpa`), being a measure of accuracy of Predicted Positives in contrast with the rate of discovery of Real Positives (`tpr`). Precision is defined in (2):

Precision = Confidence = `tpa` = tp/pp
= TP / PP = A / (A+B)     (2)

These two measures and their combinations focus only on the positive examples and predictions, although between them they capture some information about the rates and kinds of errors made. However, neither of them captures any information about how well the model handles negative cases. Recall relates only to the **+R** column and Precision only to the **+P** row. Neither of these takes into account the number of True Negatives. This also applies to their Arithmetic, Geometric and Harmonic Means: A, G and F=G$^2$/A (the F-factor or F-measure). Note that the F1-measure effectively references the True Positives to the Arithmetic Mean of Predicted Positives and Real Positives, being a constructed rate normalized to an idealized value, and expressed in this form it is known in statistics as a Proportion of Specific Agreement as it is a applied to a specific class, so applied to the Positive Class, it is PS+. It also corresponds to the set-theoretic Dice Coefficient. The Geometric Mean of Recall and Precision (G-measure) effectively normalizes TP to the Geometric Mean of Predicted Positives and Real Positives, and its Information content corresponds to the Arithmetic Mean of the Information represented by Recall and Precision.

In fact, there is in principle nothing special about the Positive case, and we can define Inverse statistics in terms of the Inverse problem in which we interchange positive and negative and are predicting the opposite case. Inverse Recall or Specificity is thus the proportion of Real Negative cases that are correctly

**Table 1. Systematic and traditional notations in a binary contingency table.** Shading indicates correct (light=green) and incorrect (dark=red) rates or counts in the contingency table.

|  | +R | −R |  |  |  | +R | −R |  |
|---|---|---|---|---|---|---|---|---|
|  | tp | fp | pp | **+P** | A | B | A+B |
|  | fn | tn | pn | **−P** | C | D | C+D |
|  | rp | rn | 1 |  | A+C | B+D | N |

Predicted Negative (3), and is also known as the True Negative Rate (`tnr`). Conversely, Inverse Precision is the proportion of Predicted Negative cases that are indeed Real Negatives (4), and can also be called True Negative Accuracy (`tna`):

Inverse Recall  =`tnr`      =`tn/rn`
                =`TN/RN`    =`D/(B+D)`   (3)
Inverse Precision =`tna`    =`tn/pn`
                =`TN/PN`    =`D/(C+D)`   (4)

The inverse of F1 is not known in AI/ML/CL/IR but is just as well known as PS+ in statistics, being the Proportion of Specific Agreement for the class of negatives, PS−. Note that where as F1 is advocated in AI/ML/CL/IR as a single measure to capture the effectiveness of a system, it still completely ignores `TN` which can vary freely without affecting the statistic. In statistics, PS+ is used in conjunction with PS− to ensure the contingencies are completely captured, and similarly Specificity (Inverse Recall) is always recorded along with Sensitivity (Recall).

Rand Accuracy explicitly takes into account the classification of negatives, and is expressible (5) both as a weighted average of Precision and Inverse Precision and as a weighted average of Recall and Inverse Recall:

Accuracy  =`tca`=`tcr`=`tp+tn`
          =`rp·tpr+rn·tnr`  =`(TP+TN)/N`
          =`pp·tpa+pn·tna`  =`(A+D)/N`   (5)
Dice = F1 =`tp/(tp+(fn+fp)/2)`
          =`A/(A+(B+C)/2)`               (6)
          =`1/(1+mean(FN,FP)/TP)`
Jaccard   =`tp/(tp+fn+fp)`=`TP/(N-TN)`
          =`A/(A+B+C)`  = `A/(N-D)`      (7)
          =`1/(1+2mean(FN,FP)/TP)`
          = F1 / (2 − F1)

As shown in (5) Rand Accuracy is effectively a *prevalence-weighted* average of Recall and Inverse Recall, as well as a *bias-weighted* average of Precision and Inverse Precision. Whilst it does take into account `TN` in the numerator, the sensitivity to bias and prevalence is an issue since these are independent variables, with prevalence varying as we apply to data sampled under different conditions, and bias being directly under the control of the system designer (e.g. as a threshold). Similarly, we can note that one of `N,FP` or `FN` is free to vary. Whilst it apparently takes into account `TN` in the numerator, the Jaccard (or Tanimoto) similarity coefficient uses it to heuristically *discount* the correct classification of negatives, but it can be written (6) independently of `FN` and `N` in a way similar to the effectively equivalent Dice or PS+ or F1 (7), or in terms of them, and so is subject to bias as `FN` or `N` is free to vary and they fail to capture contingencies fully without knowing inverse statistics too.

Each of the above also has a complementary form defining an error rate, of which some have specific names and importance: Fallout or False Positive Rate (`fpr`) are the proportion of Real Negatives that occur as Predicted Positive (ring-ins); Miss Rate or False Negative Rate (`fnr`) are the proportion of Real Positives that are Predicted Negatives (false-drops). False Positive Rate is the second of the legs on which ROC analysis is based.

Fallout    =`fpr`     =`fp/rp`
           =`FP/RP`   =`B/(B+D)`   (8)
Miss Rate  =`fnr` =`fn/rn`
           =`FN/RN`   =`C/(A+C)`   (9)

Note that FN and FP are sometimes referred to as Type I and Type II Errors, and the rates `fn` and `fp` as alpha and beta, respectively – referring to falsely rejecting or accepting a hypothesis. More correctly, these terms apply specifically to the meta-level problem discussed later of whether the precise pattern of counts (not rates) in the contingency table fit the null hypothesis of random distribution rather than reflecting the effect of some alternative hypothesis (which is not in general the one represented by **+P**→**+R** or −**P**→ −**R** or both). Note that *all* the measures discussed *individually* leave at least two degree of freedom (plus `N`) unspecified and free to control, and this leaves the door open for bias, whilst `N` is needed too for estimating significance and power.

**Prevalence, Bias, Cost & Skew**

We now turn our attention to the various forms of bias that detract from the utility of all of the above surface measures [2]. We will first note that `rp` represents the Prevalence of positive cases, `RP/N`, and is assumed to be a property of the population of interest – it may be constant, or it may vary across subpopulations, but is regarded here as not being under the control of the experimenter, and so we want a prevalence independent measure. By contrast, `pp` represents the (label) Bias of the model [3], the tendency of the model to output positive labels, `PP/N`, and is directly under the control of the experimenter, who can change the model by changing the theory or algorithm, or some parameter or threshold, to better fit the world/population being modeled. As discussed earlier, F-factor (or Dice or Jaccard) effectively references `tp` (probability or proportion of True Positives) to the Arithmetic Mean of Bias and Prevalence (6-7). A common rule of thumb, or even a characteristic of some algorithms, is to parameterize a model so that Prevalence = Bias, viz. `rp = pp`. Corollaries of this setting are Recall = Precision (= Dice but not Jaccard), Inverse Recall = Inverse Precision and Fallout = Miss Rate.

Alternate characterizations of Prevalence are in terms of Odds [4] or Skew [5], being the Class Ratio $c_s$ = `rn/rp`, recalling that by definition `rp+rn = 1` and RN+RP = N. If the distribution is highly skewed, typically there are many more negative cases than positive, this means the number of errors due to poor Inverse Recall will be much greater than the number of errors due to poor Recall. Given the cost of both False Positives and False Negatives is equal, individually, the overall component of the total cost due to False Positives (as Negatives) will be much greater at any significant level of chance performance, due to the higher Prevalence of Real Negatives.

Note that the normalized binary contingency table with unspecified margins has three degrees of freedom – setting any three non−Redundant ratios determines the rest (setting any count supplies the remaining information to recover the original table of counts with its four degrees of freedom). In particular, Recall, Inverse Recall and Prevalence, or equivalently tpr, fpr and $c_s$, suffice to determine all ratios and measures derivable from the normalized contingency table, but N is also required to determine significance. As another case of specific interest, Precision, Inverse Precision and Bias, in combination, suffice to determine all ratios or measures, although we will show later that an alternate characterization of Prevalence and Bias in terms of Evenness allows for even simpler relationships to be exposed.

We can also take into account a differential value for positives (cp) and negatives (cn) – this can be applied to errors as a cost (loss or debit) and/or to correct cases as a gain (profit or credit), and can be combined into a single Cost Ratio $c_v$= cn/cp. Note that the value and skew determined costs have similar effects, and may be multiplied to produce a single skew-like cost factor c = $c_v c_s$. Formulations of measures that are expressed using tpr, fpr and $c_s$ may be made cost-sensitive by using c = $c_v c_s$ in place of c = $c_s$, or can be made skew/cost-insensitive by using c = 1[5].

**ROC and PN Analyses**

Flach [5] highlighted the utility of ROC analysis to the Machine Learning community, and characterized the skew sensitivity of many measures in that context, utilizing the ROC format to give geometric insights into the nature of the measures and their sensitivity to skew. [6] further elaborated this analysis, extending it to the unnormalized PN variant of ROC, and targeting their analysis specifically to rule learning. We will not examine the advantages of ROC analysis here, but will briefly explain the principles and recapitulate some of the results.

ROC analysis plots the rate tpr against the rate fpr, whilst PN plots the unnormalized TP against FP. This difference in normalization only changes the scales and gradients, and we will deal only with the normalized form of ROC analysis. A perfect classifier will score in the top left hand corner (fpr=0,tpr=100%). A worst case classifier will score in the bottom right hand corner (fpr=100%,tpr=0). A random classifier would be expected to score somewhere along the positive diagonal (tpr=fpr) since the model will throw up positive and negative examples at the same rate (relative to their populations – these are Recall-like scales: tpr = Recall, 1-fpr = Inverse Recall). For the negative diagonal (tpr+c·fpr=1) corresponds to matching Bias to Prevalence for a skew of c.

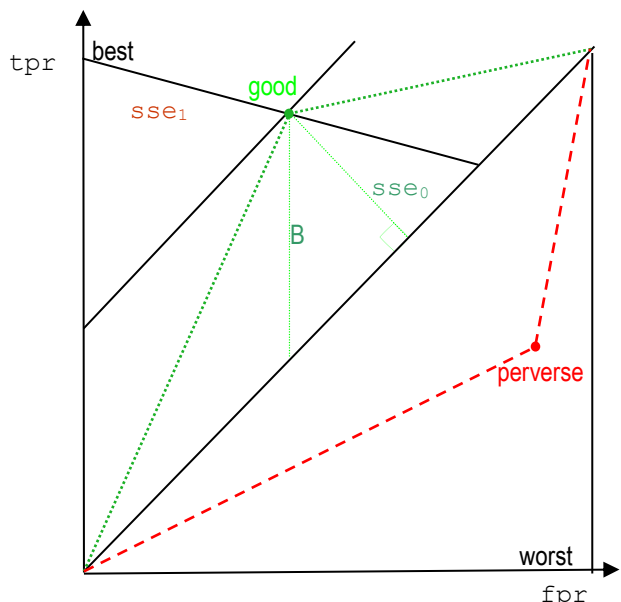

**Figure 1. Illustration of ROC Analysis.** The main diagonal represents chance with parallel isocost lines representing equal cost-performance. Points above the diagonal represent performance better than chance, those below worse than chance. For a single good (dotted=green) system, AUC is area under curve (trapezoid between green line and x=[0,1] ). The perverse (dashed=red) system shown is the same (good) system with class labels reversed.

The ROC plot allows us to compare classifiers (models and/or parameterizations) and choose the one that is closest to (0,1) and furtherest from tpr=fpr in some sense. These conditions for choosing the optimal parameterization or model are not identical, and in fact the most common condition is to minimize the area under the curve (AUC), which for a single parameterization of a model is defined by a single point and the segments connecting it to (0,0) and (1,1). For a parameterized model it will be a monotonic function consisting of a sequence of segments from (0,0) to (1,1). A particular cost model and/or accuracy measure defines an isocost gradient, which for a skew and cost insensitive model will be c=1, and hence another common approach is to choose a tangent point on the highest isocost line that touches the curve. The simple condition of choosing the point on the curve nearest the optimum point (0,1) is not commonly used, but this distance to (0,1) is given by $\sqrt{[(-\text{fpr})^2 + (1-\text{tpr})^2]}$, and minimizing this amounts to minimizing the sum of squared normalized error, $\text{fpr}^2+\text{fnr}^2$.

A ROC curve with concavities can also be locally interpolated to produce a smoothed model following the convex hull of the original ROC curve. It is even possible to locally invert across the convex hull to repair concavities, but this may overfit and thus not generalize to unseen data. Such repairs can lead to selecting an improved model, and the ROC curve can also be used to return a model to changing Prevalence and costs. The area under such a multipoint curve is thus of some value, but the optimum in practice is the area under the simple trapezoid defined by the model:

```
AUC   = (tpr-fpr+1)/2
      = (tpr+tnr)/2
      = 1 – (fpr+fnr)/2           (10)
```

For the cost and skew insensitive case, with `c=1`, maximizing AUC is thus equivalent to maximizing `tpr-fpr` or minimizing a sum of (absolute) normalized error `fpr+fnr`. The chance line corresponds to `tpr-fpr=0`, and parallel isocost lines for `c=1` have the form `tpr-fpr=k`. The highest isocost line also maximizes `tpr-fpr` and AUC so that these two approaches are equivalent. Minimizing a sum of squared normalized error, $\text{fpr}^2+\text{fnr}^2$, corresponds to a Euclidean distance minimization heuristic that is equivalent only under appropriate constraints, e.g. `fpr=fnr`, or equivalently, Bias=Prevalence, noting that all cells are non-negative by construction.

We now summarize relationships between the various candidate accuracy measures as rewritten [5,6] in terms of `tpr`, `fpr` and the skew, `c`, as well in terms of Recall, Bias and Prevalence:

```
Accuracy   = [tpr+c·(1-fpr)]/[1+c]
           = 2·Recall·Prev+1–Bias–Prev     (11)
Precision  = tpr/[tpr+c·fpr]
           = Recall·Prev/Bias              (12)
F-Measure F1 = 2·tpr/[tpr+c·fpr+1]
           = 2·Recall·Prev/[Bias+Prev]     (13)
WRacc      = 4c·[tpr-fpr]/[1+c]²
           = 4·[Recall–Bias]·Prev          (14)
```

The last measure, Weighted Relative Accuracy, was defined [7] to subtract off the component of the True Positive score that is attributable to chance and rescale to the range ±1. Note that maximizingWRacc is equivalent to maximizing AUC or `tpr-fpr` =2·AUC−1, as `c` is constant. Thus WRAcc is an unbiased accuracy measure, and the skew-insensitive form of WRAcc, with `c=1`, is precisely `tpr-fpr`. Each of the other measures (10−12) shows a bias in that it can not be maximized independent of skew, although skew-insensitive versions can be defined by setting `c=1`. The recasting of Accuracy, Precision and F-Measure in terms of Recall makes clear how all of these vary only in terms of the way they are affected by Prevalence and Bias.

Prevalence is regarded as a constant of the target condition or data set (and `c=[1-Prev]/Prev`), whilst parameterizing or selecting a model can be viewed in terms of trading off `tpr` and `fpr` as in ROC analysis, or equivalently as controlling the relative number of positive and negative predictions, namely the Bias, in order to maximize a particular accuracy measure (Recall, Precision, F-Measure, Rand Accuracy and AUC). Note that for a given Recall level, the other measures (10−13) all decrease with increasing Bias towards positive predictions.

**DeltaP, Informedness and Markedness**

Powers [4] also derived an unbiased accuracy measure to avoid the bias of Recall, Precision and Accuracy due to population Prevalence and label bias. The Bookmaker algorithm costs wins and losses in the same way a fair bookmaker would set prices based on the odds. Powers then defines the concept of Informedness which represents the 'edge' a punter has in making his bet, as evidenced and quantified by his winnings. Fair pricing based on correct odds should be zero sum – that is, guessing will leave you with nothing in the long run, whilst a punter with certain knowledge will win every time. Informedness is the probability that a punter is making an informed bet and is explained in terms of the proportion of the time the edge works out versus ends up being pure guesswork. Powers defined Bookmaker Informedness for the general, `K`-label, case, but we will defer discussion of the general case for now and present a simplified formulation of Informedness, as well as the complementary concept of Markedness.

**Definition 1**

*Informedness quantifies how informed a predictor is for the specified condition, and specifies the probability that a prediction is informed in relation to the condition (versus chance).*

**Definition 2**

*Markedness quantifies how marked a condition is for the specified predictor, and specifies the probability that a condition is marked by the predictor (versus chance).*

These definitions are aligned with the psychological and linguistic uses of the terms condition and marker. The condition represents the experimental outcome we are trying to determine by indirect means. A marker or predictor (cf. biomarker or neuromarker) represents the indicator we are using to determine the outcome. There is no implication of causality – that is something we will address later. However there are two possible directions of implication we will address now. Detection of the predictor may reliably predict the outcome, with or without the occurrence of a specific outcome condition reliably triggering the predictor.

For the binary case we have

Informedness = Recall + Inverse Recall − 1
             = tpr-fpr = 1-fnr-fpr  (15)
Markedness   = Precision + Inverse Precision − 1
             = tpa-fna = 1-fpa-fna

We noted above that maximizing AUC or the unbiased WRAcc measure effectively maximized tpr-fpr and indeed WRAcc reduced to this in the skew independent case. This is not surprising given both Powers [4] and Flach [5-7] set out to produce an unbiased measure, and the linear definition of Informedness will define a unique linear form. Note that while Informedness is a deep measure of how consistently the Predictor predicts the Outcome by combining surface measures about what proportion of Outcomes are correctly predicted, Markedness is a deep measure of how consistently the Outcome has the Predictor as a Marker by combining surface measures about what proportion of Predictions are correct.

In the Psychology literature, Markedness is known as DeltaP and is empirically a good predictor of human associative judgements – that is it seems we develop associative relationships between a predictor and an outcome when DeltaP is high, and this is true even when multiple predictors are in competition [8]. In the context of experiments on information use in syllable processing, [9] notes that Schanks [8] sees DeltaP as "the normative measure of contingency", but propose a complementary, backward, additional measure of strength of association, DeltaP' aka dichotomous Informedness. Perruchet and Peeremant [9] also note the analog of DeltaP to regression coefficient, and that the Geometric Mean of the two measures is a dichotomous form of the Pearson correlation coefficient, the Matthews' Correlation Coefficient, which is appropriate unless a continuous scale is being measured dichotomously in which case a Tetrachoric Correlation estimate would be appropriate [10,11].

**Causality, Correlation and Regression**

In a linear regression of two variables, we seek to predict one variable, y, as a linear combination of the other, x, finding a line of best fit in the sense of minimizing the sum of squared error (in y). The equation of fit has the form

y= y$_0$ + r$_x$·x            where
r$_x$= [n∑x·y-∑x·∑y]/[n∑x$^2$-∑x·∑x]           (16)

Substituting in counts from the contingency table, for the regression of predicting **+R** (1) versus-**R** (0) given **+P** (1) versus-**P** (0), we obtain this gradient of best fit (minimizing the error in the real values **R**):

r$_P$ =  [AD−BC] / [(A+B)(C+D)]
     =  A/(A+B) − C/(C+D)
     =  DeltaP = Markedness           (17)

Conversely, we can find the regression coefficient for predicting **P** from **R** (minimizing the error in the predictions **P**):

r$_R$ =  [AD−BC] / [(A+C)(B+D)]
     =  A/(A+C) − B/(B+D)
     =  DeltaP' = Informedness           (18)

Finally we see that the Matthews correlation, a contingency matrix method of calculating the Pearson product-moment correlation coefficient, ρ, is defined by

r$_G$ =[AD−BC]/√[(A+C)(B+D)(A+B)(C+D)]
     =Correlation
     =±√[Informedness·Markedness]           (19)

Given the regressions find the same line of best fit, these gradients should be reciprocal, defining a perfect Correlation of 1. However, both Informedness and Markedness are probabilities with an upper bound of 1, so perfect correlation requires perfect regression. The squared correlation is a coefficient of proportionality indicating the proportion of the variance in R that is explained by P, and is traditionally also interpreted as a probability. We can now interpret it either as the joint probability that P informs R and R marks P, given that the two directions of predictability are independent, or as the probability that the variance is (causally) explained reciprocally. The sign of the Correlation will be the same as the sign of Informedness and Markedness and indicates whether a correct or perverse usage of the information has been made – take note in interpreting the final part of (19).

Psychologists traditionally explain DeltaP in terms of causal prediction, but it is important to note that the direction of stronger prediction is not necessarily the direction of causality, and the fallacy of abductive reasoning is that the truth of A → B does not in general have any bearing on the truth of B → A.

If **Pi** is one of several independent possible causes of **R**, **Pi**→**R** is strong, but **R** →**Pi** is in general weak for any specific **Pi**. If **Pi** is one of several necessary contributing factors to **R**, **Pi**→**R** is weak for any single **Pi**, but **R** →**Pi** is strong. The directions of the implication are thus not in general dependent.

In terms of the regression to fit **R** from **P**, since there are only two correct points and two error points, and errors are calculated in the vertical (**R**) direction only, all errors contribute equally to tilting the regression down from the ideal line of fit. This Markedness regression thus provides information about the consistency of the Outcome in terms of having the Predictor as a Marker – the errors measured from the Outcome **R** relate to the failure of the Marker **P** to be present.

We can gain further insight into the nature of these regression and correlation coefficients by reducing the top and bottom of each expression to probabilities (dividing by $N^2$, noting that the original contingency counts sum to N, and the joint probabilities after reduction sum to 1). The numerator is the determinant of the contingency matrix, and common across all three coefficients, reducing to `dtp`, whilst the reduced denominator of the regression coefficients depends only on the Prevalence or Bias of the base variates. The regression coefficients, Bookmaker Informedness (B) and Markedness (M), may thus be re-expressed in terms of Precision (Prec) or Recall, along with Bias and Prevalence (Prev) or their inverses (I-):

$$
\begin{aligned}
M &= \mathtt{dtp}/\ [\text{Bias} \cdot (1-\text{Bias})]\\
  &= \mathtt{dtp}/\ [\text{pp·pn}] = \mathtt{dtp}\ /\ \text{pg}^2\\
  &= \mathtt{dtp}\ /\ \text{BiasG}^2 = \mathtt{dtp}\ /\ \text{Evenness}_\mathbf{P}\\
  &= [\text{Precision} - \text{Prevalence}]\ /\ \text{IBias} \quad (20)\\
B &= \mathtt{dtp}/\ [\text{Prevalence}\cdot(1-\text{Prevalence})]\\
  &= \mathtt{dtp}/\ [\text{rp·rn}] = \mathtt{dtp}\ /\ \text{rg}^2\\
  &= \mathtt{dtp}\ /\ \text{PrevG}^2 = \mathtt{dtp}\ /\ \text{Evenness}_\mathbf{R}\\
  &= [\text{Recall} - \text{Bias}]\ /\ \text{IPrev}\\
  &= \text{Recall} - \text{Fallout}\\
  &= \text{Recall} + \text{IRecall} - 1\\
  &= \text{Sensitivity} + \text{Specificity} - 1\\
  &= (\text{LR}-1)\cdot(1-\text{Specificity})\\
  &= (1-\text{NLR})\cdot \text{Specificity}\\
  &= (\text{LR}-1)\cdot(1-\text{NLR})\ /\ (\text{LR}-\text{NLR}) \quad (21)
\end{aligned}
$$

In the medical and behavioural sciences, the Positive Likelihood Ratio is LR=Sensitivity/[1−Specificity], and the Negative form is NLR= [1−Sensitivity]/Specificity. For non-negative B, we expect LR≥1≥NLR, with 1 as the chance case. We also express Informedness in these terms in (21).

The Matthews/Pearson correlation is expressed in reduced form as the Geometric Mean of Bookmaker Informedness and Markedness, abbreviating their product as BookMark (BM) and recalling that it is BookMark that acts as a probability-like coefficient of determination, not its root, the Geometric Mean (BookMarkG or BMG):

$$
\begin{aligned}
\text{BMG} &= \mathtt{dtp}/\ \sqrt{[\text{Prev}\cdot(1-\text{Prev})\cdot\text{Bias}\cdot(1-\text{Bias})]}\\
 &= \mathtt{dtp}\ /\ [\text{PrevG}\cdot\text{BiasG}]\\
 &= \mathtt{dtp}\ /\ \text{Evenness}_\mathbf{G}\\
 &= \sqrt{[(\text{Recall}-\text{Bias})(\text{Prec}-\text{Prev})]/(\text{IPrev}\cdot\text{IBias})} \quad (22)
\end{aligned}
$$

These equations clearly indicate how the Bookmaker coefficients of regression and correlation depend only on the proportion of True Positives and the Prevalence and Bias applicable to the respective variables. Furthermore, Prev · Bias represents the Expected proportion of True Positives (`etp`) relative to N, showing that the coefficients each represent the proportion of Delta True Positives (deviation from expectation, `dtp=tp-etp`) renormalized in different ways to give different probabilities. Equations (20-22) illustrate this, showing that these coefficients depend only on `dtp` and either Prevalence, Bias or their combination. Note that for a particular `dtp` these coefficients are minimized when the Prevalence and/or Bias are at the evenly biased 0.5 level, however in a learning or parameterization context changing the Prevalence or Bias will in general change both `tp` and `etp`, and hence can change `dtp`.

It is also worth considering further the relationship of the denominators to the Geometric Means, PrevG of Prevalence and Inverse Prevalence (IPrev = 1−Prev is Prevalence of Real Negatives) and BiasG of Bias and Inverse Bias (IBias = 1−Bias is bias to Predicted Negatives). These Geometric Means represent the Evenness of Real classes (Evenness$_\mathbf{R}$ = PrevG$^2$) and Predicted labels (Evenness$_\mathbf{P}$ = BiasG$^2$). We also introduce the concept of Global Evenness as the Geometric Mean of these two natural kinds of Evenness, Evenness$_\mathbf{G}$. From this formulation we can see that for a given relative delta of true positive prediction above expectation (`dtp`), the correlation is at minimum when predictions and outcomes are both evenly distributed (√Evenness$_\mathbf{G}$ = √Evenness$_\mathbf{R}$ = √Evenness$_\mathbf{P}$ = Prev = Bias = 0.5), and Markedness and Bookmaker are individually minimal when Bias resp. Prevalence are evenly distributed (viz. Bias resp. Prev = 0.5). This suggests that setting Learner Bias (and regularized, cost-weighted or subsampled Prevalence) to 0.5, as sometimes performed in Artificial Neural Network training is in fact inappropriate on theoretical grounds, as has Previously been shown both empirically and based on Bayesian principles – rather it is best to use Learner/Label Bias = Natural Prevalence which is in general much less than 0.5 [12].

Note that in the above equations (20-22) the denominator is always strictly positive since we have occurrences and predictions of both Positives and Negatives by earlier assumption, but we note that if in violation of this constraint we have a degenerate case in which there is nothing to predict or we make no effective prediction, then `tp=etp` and `dtp=0`, and all the above regression and correlation coefficients are defined in the limit approaching zero. Thus the coefficients are zero if and only if `dtp` is zero, and they have the same sign as `dtp` otherwise. Assuming that we are using the model the right way round, then `dtp`, B and M are non-negative, and BMG is similarly non-negative as expected. If the model is the wrong way round, then `dtp`, B, M and BMG can indicate this by expressing below chance performance, negative regressions and negative correlation, and we can reverse the sense of **P** to correct this.

The absolute value of the determinant of the contingency matrix, dp= `dtp`, in these probability formulae (20-22), also represents the sum of absolute deviations from the expectation represented by any individual cell and hence `2dp=2DP/N is` the total absolute relative error versus the null hypothesis. Additionally it has a geometric interpretation as the area of a trapezoid in PN-space, the unnormalized variant of ROC [6].

We already observed that in (normalized) ROC analysis, Informedness is twice the triangular area between a positively informed system and the chance line, and it thus corresponds to the area of the trapezoid defined by a system (assumed to perform no worse than chance), and any of its perversions (interchanging prediction labels but not the real classes, or vice-versa, so as to derive a system that performs no better than chance), and the endpoints of the chance line (the trivial cases in which the system labels all cases true or conversely all are labelled false). Such a kite-shaped area is delimited by the dotted (system) and dashed (perversion) lines in Fig. 1 (interchanging class labels), but the alternate parallelogram (interchanging prediction labels) is not shown. The Informedness of a perverted system is the negation of the Informedness of the correctly polarized system.

We now also express the Informedness and Markedness forms of DeltaP in terms of deviations from expected values along with the Harmonic mean of the marginal cardinalities of the Real classes or Predicted labels respectively, defining `DP, DELTAP, RH, PH` and related forms in terms of their `N−Relative` probabilistic forms defined as follows:

```
etp    = rp·pp; etn = rn· pn  (23)
dp     = tp − etp = dtp
       = −dtn = −(tn − etn)
deltap = dtp − dtn = 2dp         (24)
rh = 2rp·rn / [rp+rn] = rp²/ra²
ph = 2pp·pn / [pp+pn] = pp²/pa²  (25)
```

DeltaP' or Bookmaker Informedness may now be expressed in terms of `deltap` and `rh`, and DeltaP or Markedness analogously in terms of `deltap` and `ph`:

```
B = DeltaP' = [etp+dtp]/rp−[efp−dtp]/rn
            = etp/rp − efp/rn + 2dtp/rh
            = 2dp/rh = deltap/rh        (26)
M = DeltaP  = 2dp/ph = deltap/ph        (27)
```

These harmonic relationships connect directly with the previous geometric evenness terms by observing HarmonicMean = GeometricMean²/ArithmeticMean as seen in (25) and used in the alternative expressions for normalization for Evenness in (26-27). The use of HarmonicMean makes the relationship with F-measure clearer, but use of GeometricMean is generally preferred as a consistent estimate of central tendency that more accurately estimates the mode for skewed (e.g. Poisson) data bounded below by 0 and unbounded above, and as the central limit of the family of $L_p$ based averages. Viz. the Geometric ($L_0$) Mean is the Geometric Mean of the Harmonic ($L_{-1}$) and Arithmetic ($L_{+1}$) Means, with positive values of p being biased higher (toward $L_{+\infty}$=Max) and negative values of p being biased lower (toward $L_{-\infty}$=Min).

**Effect of Bias and Prev on Recall and Precision**

The final form of the equations (26-27) cancels out the common Bias and Prevalence (Prev) terms, that denormalized `tp` to `tpr` (Recall) or `tpa` (Precision). We now recast the Bookmaker Informedness and Markedness equations to show Recall and Precision as subject (28-29), in order to explore the affect of Bias and Prevalence on Recall and Precision, as well as clarify the relationship of Bookmaker and Markedness to these other ubiquitous but iniquitous measures.

Recall     = Bookmaker (1−Prevalence) + Bias
Bookmaker  = (Recall-Bias)/(1−Prevalence)    (28)
Precision  = Markedness (1-Bias) + Prevalence
Markedness = (Precision−Prevalence)/(1-Bias) (29)

Bookmaker and Markedness are unbiased estimators of above chance performance (relative to respectively the predicting conditions or the predicted markers). Equations (28-29) clearly show the nature of the bias introduced by both Label Bias and Class Prevalence. If operating at chance level, both Bookmaker and Markedness will be zero, and Recall, Precision, and derivatives such as the F-measure, will be skewed by the biases. Note that increasing Bias or decreasing Prevalence increases Recall and decreases Precision, for a constant level of unbiased performance. We can more specifically see that the regression coefficient for the prediction of Recall from Prevalence is −Informedness, and from Bias is +1, and similarly the regression coefficient for the prediction of Precision from Bias is −Markedness, and from Prevalence is +1. Using the heuristic of setting Bias = Prevalence then sets Recall = Precision = F1 and Bookmaker Informedness = Markedness = Correlation. Setting Bias = 1 (Prevalence<1) may be seen to make Precision track Prevalence with Recall = 1, whilst Prevalence = 1 (Bias<1) means Recall = Bias with Precision = 1, and under either condition no information is utilized (Bookmaker Informedness = Markedness = 0).

In summary, Recall reflects the Bias plus a discounted estimation of Informedness and Precision reflects the Prevalence plus a discounted estimation of Markedness. Given usually Prevalence << ½ and Bias << ½, their complements Inverse Prevalence >> ½ and Inverse Bias >> ½ represent substantial weighting up of the true unbiased performance in both these measures, and hence also in F1. High Bias drives

Recall up strongly and Precision down according to the strength of Informedness; high Prevalence drives Precision up and Recall down according to the strength of Markedness.

Alternately, Informedness can be viewed (21) as a renormalization of Recall after subtracting off the chance level of Recall, Bias, and Markedness (20) can be seen as a renormalization of Precision after subtracting off the chance level of Precision, Prevalence (and Flach's WRAcc, the unbiased form being equivalent to Bookmaker Informedness, was defined in this way as discussed in §2.3). Informedness can also be seen (21) as a renormalization of LR or NLR after subtracting off their chance level performance. The Kappa measure [13-16] commonly used in assessor agreement evaluation was similarly defined as a renormalization of Accuracy after subtracting off an estimate of the expected Accuracy, for Cohen Kappa being the dot product of the Biases and Prevalences, and expressible as a normalization of the discriminant of contingency, `dtp,` by the mean error rate (cf. F1; viz. Kappa is `dtp/[dtp+mean(fp,fn)]`). All three measures are invariant in the sense that they are properties of the contingency tables that remain unchanged when we flip to the Inverse problem (interchange positive and negative for both conditions and predictions). That is we observe:

Inverse Informedness = Informedness,
Inverse Markedness = Markedness,
Inverse Kappa = Kappa.

The Dual problem (interchange antecedent and consequent) reverses which condition is the predictor and the predicted condition, and hence interchanges Precision and Recall, Prevalence and Bias, as well as Markedness and Informedness. For cross-evaluator agreement, both Informedness and Markedness are meaningful although the polarity and orientation of the contingency is arbitrary. Similarly when examining causal relationships (conventionally DeltaP vs DeltaP'), it is useful to evaluate both deductive and abductive directions in determining the strength of association. For example, the connection between cloud and rain involves cloud as *one* causal antecedent of rain (but sunshowers occur occasionally), and rain as *one* causal consequent of cloud (but cloudy days aren't always wet) – only once we have identified the full causal chain can we reduce to equivalence, and lack of equivalence may be a result of unidentified causes, alternate outcomes or both.

The Perverse systems (interchanging the labels on either the predictions or the classes, but not both) have similar performance but occur below the chance line (since we have assumed strictly better than chance performance in assigning labels to the given contingency matrix).

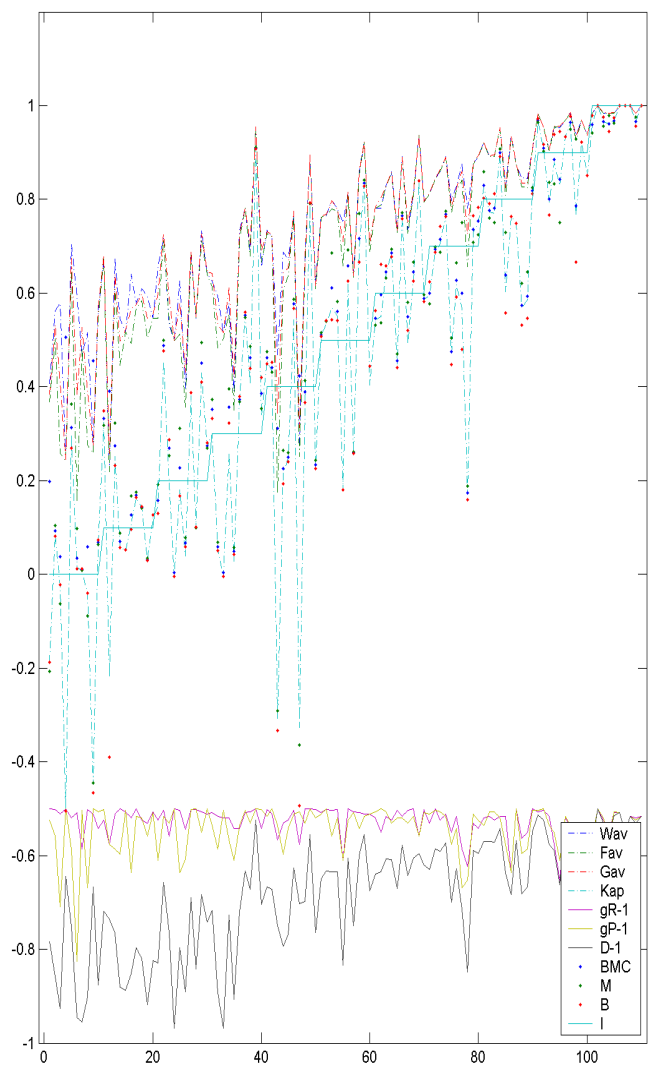

**Figure 2. Accuracy of traditional measures.**
110 Monte Carlo simulations with 11 stepped expected Informedness levels (red) with Bookmaker-estimated Informedness (red dot), Markedness (green dot) and Correlation (blue dot), and showing (dashed) Kappa versus the biased traditional measures Rank Weighted Average (Wav), Geometric Mean (Gav) and Harmonic Mean F1 (Fav). The Determinant (D) and Evenness k-th roots (gR=PrevG and gP=BiasP) are shown +1. K=4, N=128.
*(Online version has figures in colour.)*

Note that the effect of Prevalence on Accuracy, Recall and Precision has also been characterized above (§2.3) in terms of Flach's demonstration of how skew enters into their characterization in ROC analysis, and effectively assigns different costs to (False) Positives and (False) Negatives. This can be controlled for by setting the parameter `c` appropriately to reflect the desired skew and cost tradeoff, with `c=1` defining skew and cost insensitive versions. However, only Informedness (or equivalents such as DeltaP' and skew-insensitive WRAcc) precisely characterizes the probability with which a model informs the condition, and conversely only Markedness (or DeltaP) precisely characterizes the probability that a condition marks (informs) the predictor. Similarly, only the Correlation (aka Coefficient of Proportionality aka Coefficient of Determination aka Squared Matthews Correlation Coefficient) precisely characterizes the probability that condition and predictor inform/mark each other, under our dichotomous assumptions. Note the Tetrachoric Correlation is another estimate of the Pearson Correlation made under the alternate assumption of an underlying continuous variable (assumed normally distributed), and is appropriate if we instead assume that we are dichotomizing a normal continuous variable [11]. But in this article we are making the explicit assumption that we are dealing with a right/wrong dichotomy that is intrinsically discontinuous.

Although Kappa does attempt to renormalize a debiased estimate of Accuracy, and is thus much more meaningful than Recall, Precision, Accuracy, and their biased derivatives, it is intrinsically non-linear, doesn't account for error well, and retains an influence of bias, so that there does not seem that there is any situation when Kappa would be preferable to Correlation as a standard independent measure of agreement [16,13]. As we have seen, Bookmaker Informedness, Markedness and Correlation reflect the discriminant of relative contingency normalized according to different Evenness functions of the marginal Biases and Prevalences, and reflect probabilities relative to the corresponding marginal cases. However, we have seen that Kappa scales the discriminant in a way that reflects the actual error without taking into account expected error due to chance, and in effect it is really just using the discriminant to scale the actual mean error: Kappa is `dtp/[dtp+mean(fp,fn)]` = `1/[1+mean(fp,fn)/dtp]` which approximates for small error to `1-mean(fp,fn)/dtp`.

The relatively good fit of Kappa to Correlation and Informedness is illustrated in Fig. 2, along with the poor fit of the Rank Weighted Average and the Geometric and Harmonic (F-factor) means. The fit of the Evenness weighted determinant is perfect and not easily distinguishable but the separate components (Determinant and geometric means of Real Prevalences and Prediction Biases) are also shown (+1 for clarity).

**Significance and Information Gain**

The ability to calculate various probabilities from a contingency table says nothing about the significance of those numbers – is the effect real, or is it within the expected range of variation around the values expected by chance? Usually this is explored by considering deviation from the expected values (`ETP` and its relatives) implied by the marginal counts (`RP`, `PP` and relatives) – or from expected rates implied by the biases (Class Prevalence and Label Bias). In the case of Machine Learning, Data Mining, or other artificially derived models and rules, there is the further question of whether the training and parameterization of the model has set the 'correct' or 'best' Prevalence and Bias (or Cost) levels. Furthermore, should this determination be undertaken by reference to the model evaluation measures (Recall, Precision, Informedness, Markedness and their derivatives), or should the model be set to maximize the significance of the results?

This raises the question of how our measures of association and accuracy, Informedness, Markedness and Correlation, relate to standard measures of significance.

This article has been written in the context of a Prevailing methodology in Computational Linguistics and Information Retrieval that concentrates on target positive cases and ignores the negative case for the purpose of both measures of association and significance. A classic example is saying "water" can only be a noun because the system is inadequate to the task of Part of Speech identification and this boosts Recall and hence F-factor, or at least setting the Bias to nouns close to 1, and the Inverse Bias to verbs close to 0. Of course, Bookmaker will then be 0 and Markedness unstable (undefined, and very sensitive to any words that do actually get labelled verbs). We would hope that significance would also be 0 (or near zero given only a relatively small number of verb labels). We would also like to be able to calculate significance based on the positive case alone, as either the full negative information is unavailable, or it is not labelled.

Generally when dealing with contingency tables it is assumed that unused labels or unrepresented classes are dropped from the table, with corresponding reduction of degrees of freedom. For simplicity we have assumed that the margins are all non-zero, but the freedoms are there whether they are used or not, so we will not reduce them or reduce the table.

There are several schools of thought about significance testing, but all agree on the utility of calculating a p-value [19], by specifying some statistic or exact test T(X) and setting p = Prob(T(X) ≥ T(Data)). In our case, the Observed Data is summarized in a contingency table and there are a number of tests which can be used to evaluate the significance of the contingency table.

For example, Fisher's exact test calculates the proportion of contingency tables that are at least as favourable to the Prediction/Marking hypothesis, rather than the null hypothesis, and provides an accurate estimate of the significance of the entire contingency table without any constraints on the values or distribution. The log-likelihood-based $G^2$ test and Pearson's approximating $\chi^2$ tests are compared against a Chi-Squared Distribution of appropriate degree of freedom (`r`=1 for the binary contingency table given the marginal counts are known), and depend on assumptions about the distribution, and may focus only on the Predicted Positives.

$\chi^2$ captures the Total Squared Deviation relative to expectation, is here calculated only in relation to positive predictions as often only the overt prediction is considered, and the implicit prediction of negative case is ignored [17-19], noting that it sufficient to count `r`=1 cells to determine the table and make a significance estimate. However, $\chi^2$ is valid only for reasonably sized contingencies (one rule of thumb is that the expectation for the smallest cell is at least 5, and the Yates and Williams corrections will be discussed in due course [18,19]):

$\chi^2$<sub>+P</sub> = `(TP-ETP)`$^2$`/ETP+(FP-EFP)`$^2$`/EFP`
     = `DTP`$^2$`/ETP + DFP`$^2$`/EFP`
     = `2DP`$^2$`/EHP, EHP`
     = `2ETP·EFP/[ETP+EFP]`
     = `2N·dp`$^2$`/ehp, ehp`
     = `2etp·efp/[etp+efp]`
     = `2N·dp`$^2$`/[rh·pp]` = `N·dp`$^2$`/PrevG`$^2$`/Bias`
     = `N·B`$^2$`·Evenness`<sub>R</sub>`/Bias` = `N·r`$^2$<sub>P</sub>`·PrevG`$^2$`/Bias`
     ≈ `(N+PN) ·r`$^2$<sub>P</sub>`·PrevG`$^2$      (Bias → 1)
     = `(N+PN) ·B`$^2$`·Evenness`<sub>R</sub>      (30)

$G^2$ captures Total Information Gain, being N times the Average Information Gain in nats, otherwise known as Mutual Information, which however is normally expressed in bits. We will discuss this separately under the General Case. We deal with $G^2$ for positive predictions in the case of small effect, that is `dp` close to zero, showing that $G^2$ is twice as sensitive as $\chi^2$ in this range.

`G`$^2$<sub>+P</sub>`/2`=`TP·ln(TP/ETP) + FP·ln(FP/EFP)`
     =`TP·ln(1+DTP/ETP) +FP·ln(1+DFP/EFP)`
     ≈ `TP·(DTP/ETP) + FP·(DFP/EFP)`
     = `2N·dp`$^2$`/ehp`
     = `2N·dp`$^2$`/[rh·pp]`
     = `N·dp`$^2$`/PrevG`$^2$`/Bias`
     = `N·B`$^2$`·Evenness`<sub>R</sub>`/Bias`
     = `N·r`$^2$<sub>P</sub>`·PrevG`$^2$`/Bias`
     ≈ `(N+PN) ·r`$^2$<sub>P</sub>`·PrevG`$^2$      (Bias → 1)
     = `(N+PN) ·B`$^2$`·Evenness`<sub>R</sub>      (31)

In fact $\chi^2$ is notoriously unreliable for small N and small cell values, and $G^2$ is to be preferred. The Yates correction (applied only for cell values under 5) is to subtract 0.5 from the absolute `dp` value for that cell before squaring completing the calculation [17-19].

Our result (30-1) shows that $\chi^2$ and $G^2$ significance of the Informedness effect increases with `N` as expected,

but also with the square of Bookmaker, the Evenness of Prevalence (Evenness<sub>R</sub> = PrevG$^2$ = Prev·(1−Prev)) and the number of Predicted Negatives (viz. with Inverse Bias)! This is as expected. The more Informed the contingency regarding positives, the less data will be needed to reach significance. The more Biased the contingency towards positives, the less significant each positive is and the more data is needed to ensure significance. The Bias-weighted average over all Predictions (here for `K`=2 case: Positive and Negative) is simply `KN·B`$^2$`·PrevG`$^2$ which gives us an estimate of the significance without focussing on either case in particular.

$\chi^2$<sub>KB</sub>  = `2N·dtp`$^2$`/PrevG`$^2$
     = `2N·r`<sub>P</sub>$^2$ `·PrevG`$^2$
     = `2N·r`<sub>P</sub>$^2$ `·Evenness`<sub>R</sub>
     = `2N·B`$^2$`·Evenness`<sub>R</sub>      (32)

Analogous formulae can be derived for the significance of the Markedness effect for positive real classes, noting that Evenness<sub>P</sub> = BiasG$^2$ .

$\chi^2$<sub>KM</sub>  = `2N·dtp`$^2$`/BiasG`$^2$
     = `2N ·r`<sub>R</sub>$^2$ `· BiasG`$^2$
     = `2N·M`$^2$`·Evenness`<sub>P</sub>      (33)

The Geometric Mean of these two overall estimates for the full contingency table is

$\chi^2$<sub>KBM</sub> = `2N·dtp`$^2$`/PrevG·BiasG`
     = `2N·r`<sub>P</sub>`·r`<sub>R</sub> `·PrevG·BiasG`
     = `2N·r`$^2$<sub>G</sub>`·Evenness`<sub>G</sub>= `2N`$\rho^2$`·Evenness`<sub>G</sub>
     = `2N·B·M ·Evenness`<sub>G</sub>      (34)

This is simply the total Sum of Squares Deviance (SSD) accounted for by the correlation coefficient BMG (22) over the `N` data points discounted by the Global Evenness factor, being the squared Geometric Mean of all four Positive and Negative Bias and Prevalence terms (Evenness<sub>G</sub>= PrevG·BiasG). The less even the Bias and Prevalence, the more data will be required to achieve significance, the maximum evenness value of 0.25 being achieved with both even bias and even Prevalence. Note that for even bias or Prevalence, the corresponding positive and negative significance estimates match the global estimate.

When $\chi^2$<sub>+P</sub> or $G^2$<sub>+P</sub> is calculated for a specific label in a dichotomous contingency table, it has one degree of freedom for the purposes of assessment of significance. The full table also has one degree of freedom, and summing for goodness of fit over only the positive prediction label will clearly lead to a lower $\chi^2$ estimate than summing across the full table, and while summing for only the negative label will often give a similar result it will in general be different. Thus the weighted arithmetic mean calculated by $\chi^2$<sub>KB</sub> is an expected value independent of the arbitrary choice of which predictive variate is investigated. This is used to see whether a hypothesized main effect (the alternate hypothesis, HA) is borne out by a significant difference from the usual distribution (the null hypothesis, H0). Summing over the entire table (rather than averaging of labels), is used for $\chi^2$ or $G^2$ independence testing independent of any specific alternate hypothesis [21], and can be expected to achieve a $\chi^2$ estimate approximately twice that achieved by the above estimates, effectively cancelling out the Evenness term, and is thus far less conservative (viz. it is more likely to satisfy p<α):

$$\chi^2_{\mathbf{BM}} = N \cdot r^2_{\mathbf{G}} = N \cdot \rho^2 = N \cdot \varphi^2 = N \cdot B \cdot M \quad (35)$$

Note that this equates Pearson's Rho, ρ, with the Phi Correlation Coefficient, φ, which is defined in terms of the Inertia $\varphi^2=\chi^2/N$. We now have confirmed that not only does a factor of N connects the full contingency $G^2$ to Mutual Information (MI), but it also normalizes the full approximate $\chi^2$ contingency to Matthews/Pearson (=BMG=Phi) Correlation, at least for the dichotomous case. This tells us moreover, that MI and Correlation are measuring essentially the same thing, but MI and Phi do not tell us anything about the direction of the correlation, but the sign of Matthews or Pearson or BMG Correlation does (it is the Biases and Prevalences that are multiplied and squarerooted).

The individual or averaged goodness-of-fit estimates are in general much more conservative than full contingency table estimation of p by the Fisher Exact Test, but the full independence estimate can over inflate the statistic due to summation of more than there are degrees of freedom. The conservativeness has to do both with distributional assumptions of the $\chi^2$ or $G^2$ estimates that are only asymptotically valid as well as the approximative nature of $\chi^2$ in particular.

Also note that α bounds the probability of the null hypothesis, but 1-α is not a good estimate of the probabilty of any specific alternate hypothesis. Based on a Bayesian equal probability prior for the null hypothesis (H0, e.g. B=M=0 as population effect) and an unspecific one-tailed alternate hypothesis (HA, e.g. the measured B and C as true population effect), we can estimate new posterior probability estimates for Type I (H0 rejection, `Alpha(p)`) and Type II (HA rejection, `Beta(p)`) errors from the posthoc likelihood estimation [22]:

```
L(p)      =Alpha(p)/Beta(p)
          ≈ −e ^ p log(p)              (36)

Alpha(p)  = 1/[1+1/L(p)]               (37)
Beta(p)   = 1/[1+L(p)]                 (38)
```

**Confidence Intervals and Deviations**

An alternative to significance estimation is confidence estimation in the statistical rather than the data mining sense. We noted earlier that selecting the highest isocost line or maximizing AUC or Bookmaker Informedness, B, is equivalent to minimizing `fpr+fnr=(1−B)` or maximizing `tpr+tnr=(1+B)`, which maximizes the sum of normalized squared deviations of B from chance, $sse_\mathbf{B}=B^2$ (as is seen geometrically from Fig. 1). Note that this contrasts with minimizing the sum of squares distance from the optimum which minimizes the relative sum of squared normalized error of the aggregated contingency, `sse`$_\mathbf{B}$`=fpr`$^2$`+fnr`$^2$. However, an alternate definition calculating the sum of squared deviation from *optimum* is as a normalization the square of the minimum distance to the isocost of contingency, `sse`$_\mathbf{B}$`=(1−B)`$^2$.

This approach contrasts with the approach of considering the error versus a specific null hypothesis representing the expectation from margins. Normalization is to the range [0,1] like |B| and normalizes (due to similar triangles) all orientations of the distance between isocosts (Fig. 1). With these estimates the relative error is constant and the relative size of confidence intervals around the null and full hypotheses only depend on N as |B| and |1−B| are already standardized measures of deviation from null or full correlation respectively (σ/μ=1). Note however that if the *empirical* value is 0 or 1, these measures admit no error versus no information or full information resp. If the *theoretical* value is B=0, then a full ±1 error is possible, particularly in the discrete low N case where it can be equilikely and will be more likely than expected values that are fractional and thus likely to become zeros. If the theoretical value is B=1, then no variation is expected unless due to measurement error. Thus |1−B| reflects the maximum (low N) deviation in the absence of measurement error.

The standard Confidence Interval is defined in terms of the Standard Error, SE =√[SSE/(N·(N−1))] =√[sse/(N−1)]. It is usual to use a multiplier X of around X=2 as, given the central limit theorem applies and the distribution can be regarded as normal, a multiplier of 1.96 corresponds to a confidence of 95% that the true mean lies in the specified interval around the estimated mean, viz. the probability that the derived confidence interval will bound the true mean is 0.95 and the test thus corresponds approximately to a significance test with `alpha`=0.05 as the probability of rejecting a correct null hypothesis, or a power test with `beta`=0.05 as the probability of rejecting a true full or partial correlation hypothesis. A number of other distributions also approximate 95% confidence at 2SE.

We specifically reject the more traditional approach which assumes that both Prevalence and Bias are fixed, defining margins which in turn define a specific chance case rather than an isocost line representing all chance cases – we cannot assume that any solution on an isocost line has greater error than any other since all are by definition equivalent. The above approach is thus argued to be appropriate for Bookmaker and ROC statistics which are based on the isocost concept, and reflects the fact that most practical systems do not in fact preset the Bias or match it to Prevalence, and indeed Prevalences in early trials may be quite different from those in the field.

**Table 2. Binary contingency tables.** Colour coding highlights example counts of correct (light green) and incorrect (dark red) decisions with the resulting Bookmaker Informedness (B=WRacc=DeltaP'), Markedness (C=DeltaP), Matthews Correlation (C), Recall, Precision, Rand Accuracy, Harmonic Mean of Recall and Precision (F=F1), Geometric Mean of Recall and Precision (G), Cohen Kappa (κ), and χ² calculated using Bookmaker (χ²$_{+P}$), Markedness (χ²$_{+R}$) and standard (χ²) methods across the positive prediction or condition only, as well as calculated across the entire K=2 class contingency, all of which are designed to be referenced to alpha (α) according to the χ² distribution, with the latter more reliable due to taking into account all contingencies. Single-tailed threshold is shown for α =0.05.

|       | 68.0% | 32.0% |     |   |        |           |        |   |        | χ²@α=0.05 | 3.85 |          |      |
|-------|-------|-------|-----|---|--------|-----------|--------|---|--------|-----------|------|----------|------|
| 76.0% | 56    | 20    | 76  | B | 19.85% | Recall    | 82.35% | F | 77.78% | χ²$_{+P}$ | 1.13 | χ²$_{KB}$  | 1.72 |
| 24.0% | 12    | 12    | 24  | M | 23.68% | Precision | 73.68% | G | 77.90% | χ²$_{+R}$ | 1.61 | χ²$_{KM}$  | 2.05 |
|       | 68    | 32    | 100 | C | 21.68% | Rand Acc  | 68.00% | κ | 21.26% | χ²        | 1.13 | χ²$_{KBM}$ | 1.87 |

|       | 60.0% | 40.0% |     |   |        |           |        |   |        | χ²@α=0.05 | 3.85 |          |      |
|-------|-------|-------|-----|---|--------|-----------|--------|---|--------|-----------|------|----------|------|
| 42.0% | 30    | 12    | 42  | B | 20.00% | Recall    | 50.00% | F | 58.82% | χ²$_{+P}$ | 2.29 | χ²$_{KB}$  | 1.92 |
| 58.0% | 30    | 28    | 58  | M | 19.70% | Precision | 71.43% | G | 59.76% | χ²$_{+R}$ | 2.22 | χ²$_{KM}$  | 1.89 |
|       | 60    | 40    | 100 | C | 19.85% | Rand Acc  | 58.00% | κ | 18.60% | χ²        | 2.29 | χ²$_{KBM}$ | 1.91 |

The specific estimate of sse that we present for `alpha`, the probability of the current estimate for B occurring if the true Informedness is B=0, is $\sqrt{sse_{B0}}=|1-B|=1$, which is appropriate for testing the null hypothesis, and thus for defining unconventional error bars on B=0. Conversely, $\sqrt{sse_{B2}}=|B|=0$, is appropriate for testing deviation from the full hypothesis in the absence of measurement error, whilst $\sqrt{sse_{B2}}=|B|=1$ conservatively allows for full range measurement error, and thus defines unconventional error bars on B=M=C=1.

In view of the fact that there is confusion between the use of `beta` in relation to a specific full dependency hypothesis, B=1 as we have just considered, and the conventional definition of an arbitrary and unspecific alternate contingent hypothesis, B≠0, we designate the probability of incorrectly excluding the full hypothesis by `gamma`, and propose three possible related kinds of correction for the $\sqrt{sse}$ for `beta`: some kind of mean of |B| and |1−B| (the unweighted arithmetic mean is 1/2, the geometric mean is less conservative and the harmonic mean least conservative), the maximum or minimum (actually a special case of the last, the maximum being conservative and the minimum too low an underestimate in general), or an asymmetric interval that has one value on the null side and another on the full side (a parameterized special case of the last that corresponds to percentile-based usages like box plots, being more appropriate to distributions that cannot be assumed to be symmetric).

The $\sqrt{sse}$ means may be weighted or unweighted and in particular a self-weighted arithmetic mean gives our recommended definition, $\sqrt{sse_{B1}}=1-2|B|+2B^2$, whilst an unweighted geometric mean gives $\sqrt{sse_{B1}}=\sqrt{[|B|-B^2]}$ and an unweighted harmonic mean gives $\sqrt{sse_{B1}}=|B|-B^2$. All of these are symmetric, with the weighted arithmetic mean giving a minimum of 0.5 at B=±0.5 and a maximum of 1 at both B=0 and B=±1, contrasting maximally with $sse_{B0}$ and $sse_{B2}$ resp in these neighbourhoods, whilst the unweighted harmonic and geometric means having their minimum of 0 at both B=0 and B=±1, acting like $sse_{B0}$ and $sse_{B2}$ resp in these neighbourhoods (which there evidence zero variance around their assumed true values). The minimum at B=±0.5 for the geometric mean is 0.5 and for the harmonic mean, 0.25.

For this probabilistic |B| range, the weighted arithmetic mean is never less than the arithmetic mean and the geometric mean is never more than the arithmetic mean. These relations demonstrate the complementary nature of the weighted/arithmetic and unweighted geometric means. The maxima at the extremes is arguably more appropriate in relation to power as intermediate results should calculate squared deviations from a strictly intermediate expectation based on the theoretical distribution, and will thus be smaller on average if the theoretical hypothesis holds, whilst providing emphasized differentiation when near the null or full hypothesis. The minima of 0 at the extremes are not very appropriate in relation to significance versus the null hypothesis due the expectation of a normal distribution, but its power dual versus the full hypothesis is appropriately a minimum as perfect correlation admits no error distribution. Based on Monte Carlo simulations, we have observed that setting $sse_{B1}=\sqrt{sse_{B2}}=1-|B|$ as per the usual convention is appropriately conservative on the upside but a little broad on the downside, whilst the weighted arithmetic mean, $\sqrt{sse_{B1}}=1-2|B|+2B^2$, is sufficiently conservative on the downside, but unnecessarily conservative for high B.

Note that these two-tailed ranges are valid for Bookmaker Informedness and Markedness that can go positive or negative, but a one tailed test would be appropriate for unsigned statistics or where a particular direction of prediction is assumed as we have for our contingency tables. In these cases a smaller multiplier of 1.65 would suffice, however the convention is to use the overlapping of the confidence bars around the various hypotheses (although usually the null is not explicitly represented).

Thus for any two hypotheses (including the null hypothesis, or one from a different contingency table or other experiment deriving from a different theory or system) the traditional approach of checking that 1.95SE (or 2SE) error bars don't overlap is rather conservative (it is enough for the value to be outside the range for a two-sided test), whilst checking overlap of 1SE error bars is usually insufficiently conservative given that the upper represents `beta<alpha`. Where it is expected that one will be better than the other, a 1.65SE error bar including the mean for the other hypothesis is enough to indicate significance (or power=1−beta) corresponding to `alpha` (or `beta`) as desired.

The traditional calculation of error bars based on Sum of Squared Error is closely related to the calculation of Chi-Squared significance based on Total Squared Deviation, and like it are not reliable when the assumptions of normality are not approximated, and in particular when the conditions for the central limit theorem are not satisfied (e.g. N<12 or cell-count<5). They are not appropriate for application to probabilistic measures of association or error. This is captured by the meeting of the X=2 error bars for the full ($sse_{B2}$) and null ($sse_{B0}$) hypotheses at N=16 (expected count of only 4 per cell).

Here we have considered only the dichotomous case but discuss confidence intervals further below, in relation to the general case.

**SIMPLE EXAMPLES**

Bookmaker Informedness has been defined as the Probability of an informed decision, and we have shown identity with DeltaP' and WRAcc, and the close relationship (10, 15) with ROC AUC. A system that makes an informed (correct) decision for a target condition with probability B, and guesses the remainder of the time, will exhibit a Bookmaker Informedness (DeltaP') of B and a Recall of B·(1−Prev) + Bias. Conversely a proposed marker which is marked (correctly) for a target condition with probability M, and according to chance the remainder of the time, will exhibit a Markedness (DeltaP) of M and a Precision of M·(1-Bias) + Prev. Precision and Recall are thus biased by Prevalence and Bias, and variation of system parameters can make them rise or fall independently of Informedness and Markedness.

Accuracy is similarly dependent on Prevalence and Bias:

2·(B·(1−Prev)·Prev+Bias·Prev)+1− (Bias+Prev),

and Kappa has an additional problem of non-linearity due to its complex denominator:

B·(1−Prev)·Prev / (1−Bias·Prev−(Bias+Prev)/2).

It is thus useful to illustrate how each of these other measures can run counter to an improvement in overall system performance as captured by Informedness. For the examples in Table 2 (for N=100) all the other measure rise, some quite considerably, but Bookmaker actually falls. Table 2 also illustrates the usage of the Bookmaker and Markedness variants of the $\chi^2$ statistic versus the standard formulation for the positive case, showing also the full `K` class contingency version (for `K`=2 in this case).

Note that under the distributional and approximative assumptions for $\chi^2$ neither of these contingencies differ sufficiently from chance at N=100 to be significant to the 0.05 level due to the low Informedness Markedness and Correlation, however doubling the performance of the system would suffice to achieve significance at N=100 given the Evenness specified by the Prevalences and/or Biases). Moreover, even at the current performance levels the Inverse (Negative) and Dual (Marking) Problems show higher $\chi^2$ significance, approaching the 0.05 level in some instances (and far exceeding it for the Inverse Dual). The KB variant gives a single conservative significance level for the entire table, sensitive only to the direction of proposed implication, and is thus to be preferred over the standard versions that depend on choice of condition.

Incidentally, the Fisher Exact Test shows significance to the 0.05 level for both the examples in Table 2. This corresponds to an assumption of a hypergeometric distribution rather than normality – viz. all assignments of events to cells are assumed to be equally likely given the marginal constraints (Bias and Prevalence). However it is in appropriate given the Bias and Prevalence are not specified by the experimenter *in advance of the experiment* as is assumed by the conditions of this test. This has also been demonstrated empirically through Monte Carlo simulation as discussed later. See [22] for a comprehensive discussion on issues with significance testing, as well as Monte Carlo simulations.

**PRACTICAL CONSIDERATIONS**

If we have a fixed size dataset, then it is arguably sufficient to maximize the determinant of the unnormalized contingency matrix, `DT`. However this is not comparable across datasets of different sizes, and we thus need to normalize for `N`, and hence consider the determinant of the normalized contingency matrix,

`dt`. However, this value is still influenced by both Bias and Prevalence.

In the case where two evaluators or systems are being compared with no a priori preference, the Correlation gives the correct normalization by their respective Biases, and is to be preferred to Kappa.

In the case where an unimpeachable Gold Standard is employed for evaluation of a system, the appropriate normalization is for Prevalence or Evenness of the real gold standard values, giving Informedness. Since this is constant, optimizing Informedness and optimizing `dt` are equivalent.

More generally, we can look not only at what proposed solution best solves a problem, by comparing Informedness, but which problem is most usefully solved by a proposed system. In a medical context, for example, it is usual to come up with potentially useful medications or tests, and then explore their effectiveness across a wide range of complaints. In this case Markedness may be appropriate for the comparison of performance across different conditions.

Recall and Informedness, as biased and unbiased variants of the same measure, are appropriate for testing effectiveness relative to a set of conditions, and the importance of Recall is being increasingly recognized as having an important role in matching human performance, for example in Word Alignment for Machine Translation [1]. Precision and Markedness, as biased and unbiased variants of the same measure, are appropriate for testing effectiveness relative to a set of predictions. This is particularly appropriate where we do not have an appropriate gold standard giving correct labels for every case, and is the primary measure used in Information Retrieval for this reason, as we cannot know the full set of relevant documents for a query and thus cannot calculate Recall.

However, in this latter case of an incompletely characterized test set, we do not have a fully specified contingency matrix and cannot apply any of the other measures we have introduced. Rather, whether for Information Retrieval or Medical Trials, it is assumed that a test set is developed in which all real labels are reliably (but not necessarily perfectly) assigned. Note that in some domains, labels are assigned reflecting different levels of assurance, but this has lead to further confusion in relation to possible measures and the effectiveness of the techniques evaluated [1]. In Information Retrieval, the labelling of a subset of relevant documents selected by an initial collection of systems can lead to relevant documents being labelled as irrelevant because they were missed by the first generation systems – so for example systems are actually penalized for improvements that lead to discovery of relevant documents that do not contain all specified query words. Thus here too, it is important to develop test sets that of appropriate size, fully labelled, and appropriate for the correct application of both Informedness and Markedness, as unbiased versions of Recall and Precision.

This Information Retrieval paradigm indeed provides a good example for the understanding of the Informedness and Markedness measures. Not only can documents retrieved be assessed in terms of prediction of relevance labels for a query using Informedness, but queries can be assessed in terms of their appropriateness for the desired documents using Markedness, and the different kinds of search tasks can be evaluated with the combination of the two measures. The standard Information Retrieval mantra that we do not need to find *all* relevant documents (so that Recall or Informedness is not so relevant) applies only where there are huge numbers of documents containing the required information and a small number can be expected to provide that information with confidence. However another kind of Document Retrieval task involves a specific and rather small set of documents for which we need to be confident that all or most of them have been found (and so Recall or Informedness are especially relevant). This is quite typical of literature review in a specialized area, and may be complicated by new developments being presented in quite different forms by researchers who are coming at it from different directions, if not different disciplinary backgrounds.

### THE GENERAL CASE

So far we have examined only the binary case with dichotomous Positive versus Negative classes and labels.

It is beyond the scope of this article to consider the continuous or multi-valued cases, although the Matthews Correlation is a discretization of the Pearson Correlation with its continuous-valued assumption, and the Spearman Rank Correlation is an alternate form applicable to arbitrary discrete value (Likert) scales, and Tetrachoric Correlation is available to estimate the correlation of an underlying continuous scale [11]. If continuous measures corresponding to Informedness and Markedness are required due to the canonical nature of one of the scales, the corresponding Regression Coefficients are available.

It is however, useful in concluding this article to consider briefly the generalization to the multi-class case, and we will assume that both real classes and predicted classes are categorized with $K$ labels, and again we will assume that each class is non-empty unless explicitly allowed (this is because Precision is ill-defined where there are no predictions of a label, and Recall is ill-defined where there are no members of a class).

**Generalization of Association**

Powers [4] derives Bookmaker Informedness (41) analogously to Mutual Information & Conditional Entropy (39-40) as a pointwise average across the contingency cells, expressed in terms of label probabilities $P_\mathbf{P}(l)$, where $P_\mathbf{P}(l)$ is the probability of Prediction $l$, and label-conditioned class probabilities $P_\mathbf{R}(c|l)$, where $P_\mathbf{R}(c|l)$ is the probability that the Prediction labeled $l$ is actually of Real class $c$, and in particular $P_\mathbf{R}(l|l)$ = Precision($l$), and where we use the delta functions as mathematical shorthands for Boolean expressions interpreted algorithmically as in C, with true expressions taking the value 1 and false expressions 0, so that $\delta_{|c-l|} \equiv (c = l)$ represents a Dirac measure (limit as $\delta \rightarrow 0$); $\partial_{|c-l|} \equiv (c \neq l)$ represents its logical complement (1 if $c \neq l$ and 0 if $c = l$)).

$$MI(\mathbf{R}||\mathbf{P}) = \sum_l P_\mathbf{P}(l) \sum_c P_\mathbf{R}(c|l) [-\log(P_\mathbf{R}(c|l))/P_\mathbf{R}(c)] \quad (39)$$

$$H(\mathbf{R}|\mathbf{P}) = \sum_l P_\mathbf{P}(l) \sum_c P_\mathbf{R}(c|l) [-\log(P_\mathbf{R}(c|l))] \quad (40)$$

$$B(\mathbf{R}|\mathbf{P}) = \sum_l P_\mathbf{P}(l) \sum_c P_\mathbf{R}(c|l) [P_\mathbf{P}(l)/(P_\mathbf{R}(l) - \partial_{|c-l|})] \quad (41)$$

We now define a binary dichotomy for each label $l$ with $l$ and the corresponding $c$ as the Positive cases (and all other labels/classes grouped as the Negative case). We next denote its Prevalence Prev($l$) and its dichotomous Bookmaker Informedness B($l$), and so can simplify (41) to

$$B(\mathbf{R}|\mathbf{P}) = \sum_l \text{Prev}(l) \, B(l) \quad (42)$$

Analogously we define dichotomous Bias($c$) and Markedness($c$) and derive

$$M(\mathbf{P}|\mathbf{R}) = \sum_c \text{Bias}(c) \, M(c) \quad (43)$$

These formulations remain consistent with the definition of Informedness as the probability of an informed decision versus chance, and Markedness as its dual. The Geometric Mean of multi-class Informedness and Markedness would appear to give us a new definition of Correlation, whose square provides a well defined Coefficient of Determination. Recall that the dichotomous forms of Markedness (20) and Informedness (21) have the determinant of the contingency matrix as common numerators, and have denominators that relate only to the margins, to Prevalence and Bias respectively. Correlation, Markedness and Informedness are thus equal when Prevalence = Bias. The dichotomous Correlation Coefficient would thus appear to have three factors, a common factor across Markedness and Informedness, representing their conditional dependence, and factors representing Evenness of Bias (cancelled in Markedness) and Evenness of Prevalence (cancelled in Informedness), each representing a marginal independence.

In fact, Bookmaker Informedness can be driven arbitrarily close to 0 whilst Markedness is driven arbitrarily close to 1, demonstrating their independence – in this case Recall and Precision will be driven to or close to 1. The arbitrarily close hedge relates to our assumption that all predicted and real classes are non-empty, although appropriate limits could be defined to deal with the divide by zero problems associated with these extreme cases. Technically, Informedness and Markedness are conditionally independent – once the determinant numerator is fixed, their values depend only on their respective marginal denominators which can vary independently. To the extent that they are independent, the Coefficient of Determination acts as the joint probability of mutual determination, but to the extent that they are dependent, the Correlation Coefficient itself acts as the joint probability of mutual determination.

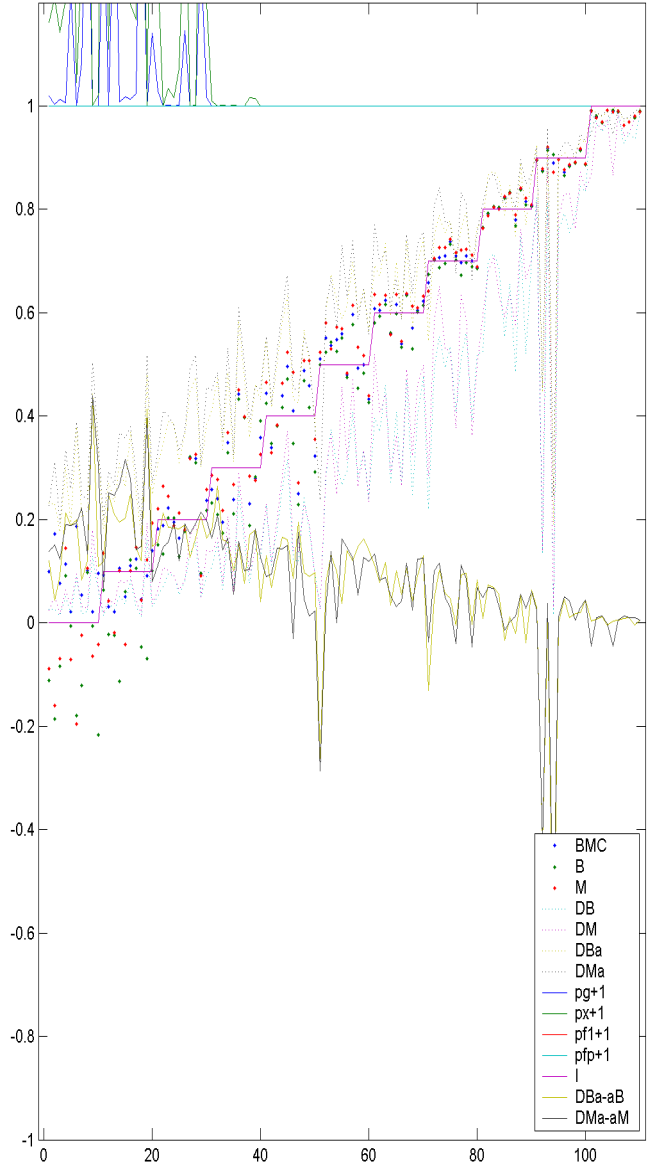

**Figure 3. Determinant-based estimates of correlation.**
110 Monte Carlo simulations with 11 stepped expected Informedness levels (red line) with Bookmaker-estimated Informedness (red dots), Markedness (green dot) and Correlation (blue dot), with significance (p+1) calculated using $G^2$, $X^2$, and Fisher estimates, and Correlation estimates calculated from the Determinant of Contingency using two different exponents, $2/K$ (DB & DM) and $1/[3K-2]$ (DBa and DMa). The difference between the estimates is also shown.
Here K=4, N=128, X=1.96, $\alpha=\beta=0.05$.

These conditions carry over to the definition of Correlation in the multi-class case as the Geometric Mean of Markedness and Informedness – once all numerators are fixed, the denominators demonstrate marginal independence.

We now reformulate the Informedness and Markedness measures in terms of the Determinant of the Contingency and Evenness, generalizing (20-22). In particular, we note that the definition of Evenness in terms of the Geometric Mean or product of biases or Prevalences is consistent with the formulation in terms of the determinants `DET` and `det` (generalizing dichotomous `DP=DTP` and `dp=dtp`) and their geometric interpretation as the area of a parallelogram in PN-space and its normalization to ROC-space by the product of Prevalences, giving Informedness, or conversely normalization to Markedness by the product of biases. The generalization of `DET` to a volume in high dimensional PN-space and `det` to its normalization by product of Prevalences or biases, is sufficient to guarantee generalization of (20-22) to `K` classes by reducing from `K`D to SSD so that BMG has the form of a coefficient of proportionality of variance:

$$M \approx [\texttt{det} / \text{BiasG}^K]^{2/K}$$
$$= \texttt{det}^{2/K} / \text{Evenness}_{P+} \quad (44)$$
$$B \approx [\texttt{det} / \text{PrevG}^K]^{2/K}$$
$$= \texttt{det}^{2/K} / \text{Evenness}_{R+} \quad (45)$$
$$\text{BMG} \approx \texttt{det}^{2/K} / [\text{PrevG} \cdot \text{BiasG}]$$
$$= \texttt{det}^{2/K} / \text{Evenness}_{G+} \quad (46)$$

We have marked the Evenness terms in these equations with a trailing plus to distinguish them from other usages, and their definitions are clear from comparison of the denominators. Note that the Evenness terms for the generalized regressions (44-45) are not Arithmetic Means but have the form of Geometric Means. Furthermore, the dichotomous case emerges for $K=2$ as expected. Empirically (Fig. 3), this generalization matches well near B=0 or B=1, but fares less well in between the extremes, suggesting a mismatched exponent in the heuristic conversion of K dimensions to 2. Here we set up the Monte Carlo simulation as follows: we define the diagonal of a random perfect performance contingency table with expected N entries using a random uniform distribution, we define a random chance level contingency table setting margins independently using a random binormal distribution, then distributing randomly across cells around their expected values, we combine the two (perfect and chance) random contingency tables with respective weights I and (1-I), and finally increment or decrement cells randomly to achieve cardinality N which is the expected number but is not constrained by the process for generating the random (perfect and chance) matrices. This procedure was used to ensure Informedness and Markedness estimates retain a level of independence; otherwise they tend to correlate very highly with overly uniform margins for higher K and lower N (conditional independence is lost once the margins are specified) and in particular Informedness, Markedness, Correlation and Kappa would always agree perfectly for either I=1 or perfectly uniform margins. Note this use of Informedness to define a target probability of an informed decision followed by random inclusion or deletion of cases when there is a mismatch versus the expected number of instances N – the preset Informedness level is thus not a fixed preset Informedness but a target level that permits jitter around that level, and in particular will be an overestimate for the step I=1 (no negative counts possible) which can be detected by excess deviation beyond the set Confidence Intervals for high Informedness steps.

In Fig. 3 we therefore show and compare an alternate exponent of $1/(3K-2)$ rather than the exponent of $2/K$ shown in (44 to 45). This also reduces to 1 and hence the expected exact correspondence for $K=2$. This suggests that what is important is not just the number of dimensions, but the also the number of marginal degrees of freedom: $K+2(K-1)$, but although it matches well for high degrees of association it shows similar error at low informedness. The precise relationship between Determinant and Correlation, Informedness and Markedness for the general case remains a matter for further investigation. We however continue with the use of the approximation based on $2/K$.

The Evenness$_R$ (Prev.IPrev) concept corresponds to the concept of Odds (IPrev/Prev), where Prev+IPrev=1, and Powers [4] shows that (multi-class) Bookmaker Informedness corresponds to the expected return per bet made with a fair Bookmaker (hence the name). From the perspective of a given bet (prediction), the return increases as the probability of winning decreases, which means that an increase in the number of other winners can increase the return for a bet on a given horse (predicting a particular class) through changing the Prevalences and thus Evenness$_R$ and the Odds. The overall return can thus increase irrespective of the success of bets in relation to those new wins. In practice, we normally assume that we are making our predictions on the basis of fixed (but not necessarily known) Prevalences which may be estimated a priori (from past data) or post hoc (from the experimental data itself), and for our purposes are assumed to be estimated from the contingency table.

**Generalization of Significance**

In relation to Significance, the single class $\chi^2_{+P}$ and $G^2_{+P}$ definitions both can be formulated in terms of cell counts and a function of ratios, and would normally be summed over at least $(K-1)^2$ cells of a $K$-class contingency table with $(K-1)^2$ degrees of freedom to produce a statistic for the table as a whole. However, these statistics are not independent of which variables are selected for evaluation or summation, and the p-values obtained are thus quite misleading, and for highly skewed distributions (in terms of Bias or Prevalence) can be outlandishly incorrect. If we sum log-likelihood (31) over all $K^2$ cells we get $N \cdot MI(\mathbf{R}||\mathbf{P})$ which is invariant over Inverses and Duals.

The analogous Prevalence-weighted multi-class statistic generalized from the Bookmaker Informedness form of the Significance statistic, and the Bias-weighted statistic generalized from the Markedness form, extend Eqns 32-34 to the $K>2$ case by probability-weighted summation (this is a weighted Arithmetic Mean of the individual cases targeted to $r=K-1$ degree of freedom):

$$\chi^2_{KB} = KN \cdot B^2 \cdot \text{Evenness}_{R-} \quad (47)$$
$$\chi^2_{KM} = KN \cdot M^2 \cdot \text{Evenness}_{P-} \quad (48)$$
$$\chi^2_{KBM} = KN \cdot B \cdot M \cdot \text{Evenness}_{G-} \quad (49)$$

For $K=2$ and $r=1$, the Evenness terms were the product of two complementary Prevalence or Bias terms in both the Bookmaker derivations and the Significance Derivations, and (30) derived a single multiplicative Evenness factor from a squared Evenness factor in the numerator deriving from $\mathtt{dtp}^2$, and a single Evenness factor in the denominator. We will discuss both these Evenness terms in the a later section. We have marked the Evenness terms in (47-49) with a trailing minus to distinguish them from forms used in (20-22,44-46).

One specific issue with the goodness-of-fit approach applied to $K$-class contingency tables relates to the up to $(K-1)^2$ degrees of freedom, which we focus on now. The assumption of independence of the counts in $(K-1)^2$ of the cells is appropriate for testing the null hypothesis, H0, and the calculation versus $\mathtt{alpha}$, but is patently not the case when the cells are generated by $K$ condition variables and $K$ prediction variables that mirror them. Thus a correction is in order for the calculation of beta for some specific alternate hypothesis $H_A$ or to examine the significance of the difference between two specific hypotheses $H_A$ and $H_B$ which may have some lesser degree of difference.

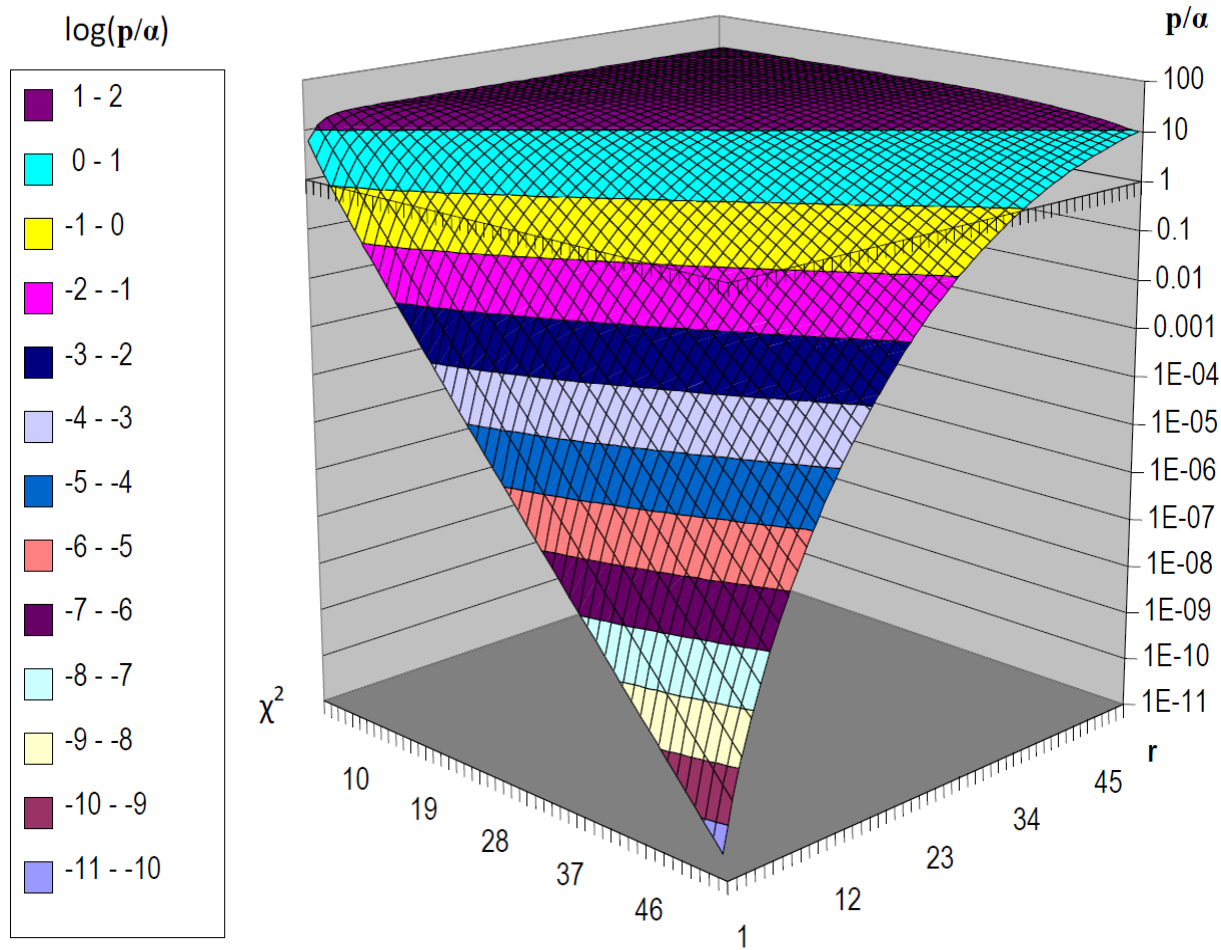

**Figure 4. Chi-squared against degrees of freedom cumulative density isocontours**
(relative to α = 0.05: cyan/yellow boundary of p/α=1=1E0)

Whilst many corrections are possible, in this case correcting the degrees of freedom directly seems appropriate and whilst using $r = (K-1)^2$ degrees of freedom is appropriate for `alpha`, using $r = K-1$ degrees of freedom is suggested for `beta` under the conditions where significance is worth testing, given the association (mirroring) between the variables is almost complete. In testing against `beta`, as a threshold on the probability that a specific alternate hypothesis of the tested association being valid should be rejected. The difference in a $\chi^2$ statistic between two systems ($r = K-1$) can thus be tested for significance as part of comparing two systems (the Correlation-based statistics are recommended in this case). The approach can also compare a system against a model with specified Informedness (or Markedness). Two special cases are relevant here, H0, the null hypothesis corresponding to null Informedness (B = 0: testing `alpha` with $r = (K-1)2$), and H1, the full hypothesis corresponding to full Informedness (B = 1: testing `beta` with $r = K-1$).

Equations 47-49 are proposed for interpretation under $r = K-1$ degrees of freedom (plus noise) and are hypothesized to be more accurate for investigating the probability of the alternate hypothesis in question, HA (`beta`).

Equations 50-52 are derived by summing over the $(K-1)$ complements of each class and label before applying the Prevalence or bias weighted sum across all predictions and conditions. These measures are thus applicable for interpretation under $r = (K-1)2$ degrees of freedom (plus biases) and are theoretically more accurate for estimating the probability of the null hypothesis H0 (`alpha`). In practice, the difference should always be slight (as the cumulative density function of the gamma distribution $\chi^2$ is locally near linear in r – see Fig. 4) reflecting the usual assumption that `alpha` and `beta` may be calculated from the same distribution. Note that there is no difference in either the formulae nor $r$ when K=2.

$$\chi^2_{XB} = K(K-1) \cdot N \cdot B^2 \cdot \text{Evenness}_{R\_} \quad (50)$$

$$\chi^2_{XBM} = K(K-1) \cdot N \cdot B \cdot M \cdot \text{Evenness}_{G\_} \quad (52)$$

Equations 53-55 are applicable to naïve unweighted summation over the entire contingency table, but also correspond to the independence test with $r = (K-1)^2$ degrees of freedom, as well as slightly underestimating but asymptotically approximating the case where Evenness is maximum in (50-52) at $1/K^2$. When the contingency table is uneven, Evenness factors will be lower and a more conservative p-value will result from (50-52), whilst summing naively across all cells (53-55) they can lead to inflated statistics and underestimated p-values. However, they are the equations that correspond to common usage of the $\chi^2$ and $G^2$ statistics as well as giving rise implicitly to Cramer's V = $[\chi^2/N(K-1)]^{1/2}$ as the corresponding estimate of the Pearson correlation coefficient, $\rho$, so that Cramer's V is thus also likely to be inflated as an estimate of association where Evenness is low. We however, note these, consistent with the usual conventions, as our definitions of the conventional forms of the $\chi^2$ statistics applied to the multiclass generalizations of the Bookmaker accuracy/association measures:

$$\chi^2_B = (K-1) \cdot N \cdot B^2 \quad (53)$$
$$\chi^2_M = (K-1) \cdot N \cdot M^2 \quad (54)$$
$$\chi^2_{BM} = (K-1) \cdot N \cdot B \cdot M \quad (55)$$

Note that Cramer's V calculated from standard full contingency $\chi^2$ and $G^2$ estimates tends vastly overestimate the level of association as measured by Bookmaker and Markedness or constructed empirically. It is also important to note that the full matrix significance estimates (and hence Cramer's V and similar estimates from these $\chi^2$ statistics) are independent of the permutations of predicted labels (or real classes) assigned to the contingency tables, and that in order to give such an independent estimate using the above family of Bookmaker statistics, it is essential that the optimal assignment of labels is made – perverse solutions with suboptimal allocations of labels will underestimate the significance of the contingency table as they clearly do take into account what one is trying to demonstrate and how well we are achieving that goal.

The empirical observation concerning Cramer's V suggests that the strict probabilistic interpretation of the multiclass generalized Informedness and Markedness measures (probability of an informed or marked decision), is not reflected by the traditional correlation measures, the squared correlation being a coefficient of proportionate determination of variance and that outside of the 2D case where they match up with BMG, we do not know how to interpret them as a probability. However, we also note that Informedness and Markedness tend to correlate and are at most conditionally independent (given any one cell, e.g given `tp`), so that their product cannot necessarily be interpreted as a joint probability (they are conditionally dependent given a margin, viz. prevalence `rp` or bias `pp`: specifying one of B or M now constrains the other; setting bias=prevalence, as a common heuristic learning constraint, maximizes correlation at BMG=B=M).

We note further that we have not considered a tetrachoric correlation, which estimates the regression of assumed underlying continuous variables to allow calculation of their Pearson Correlation.

**Sketch Proof of General Chi-squared Test**

The traditional $\chi^2$ statistic sums over a number of terms specified by $r$ degrees of freedom, stopping once dependency emerges. The $G^2$ statistic derives from a log-likelihood analysis which is also approximated, but less reliably, by the $\chi^2$ statistic. In both cases, the variates are assumed to be asymptotically normal and are expected to be normalized to mean $\mu=0$, standard deviation $\sigma=1$, and both the Pearson and Matthews correlation and the $\chi^2$ and $G^2$ significance statistics implicitly perform such a normalization. However, this leads to significance statistics that vary according to which term is in focus if we sum over `r` rather than $K^2$. In the binary dichotomous case, it makes sense to sum over only the condition of primary focus, but in the general case it involves leaving out one case (label and class). By the Central Limit Theorem, summing over $(K-1)^2$ such independent z-scores gives us a normal distribution with $\sigma=(K-1)$.

We define a single case $\chi^2_{+l\mathbf{P}}$ from the $\chi^2_{+\mathbf{P}}$ (30) calculated for label $l$ = class $c$ as the positive dichotomous case. We next sum over these for all labels other than our target $c$ to get a $(K-1)^2$ degree of freedom estimate $\chi^2_{-l\mathbf{XP}}$ given by

$$\chi^2_{-l\mathbf{XP}} = \sum_{c \neq l} \chi^2_{+l\mathbf{P}} = \sum_c \chi^2_{+c\mathbf{P}} - \chi^2_{+l\mathbf{P}} \quad (56)$$

We then perform a Bias($l$) weighted sum over $\chi^2_{-l\mathbf{XP}}$ to achieve our label independent $(K-1)^2$ degree of freedom estimate $\chi^2_{\mathbf{XB}}$ as follows (substituting from equation 30 then 39):

$$\begin{aligned}\chi^2_{\mathbf{XB}} &= \sum_l \text{Bias}(l) \cdot [N \cdot B^2 \cdot \text{Evenness}_{\mathbf{R}}(l) / \text{Bias}(l) - \chi^2_{+l\mathbf{P}}] \\ &= K \cdot \chi^2_{\mathbf{KB}} - \chi^2_{\mathbf{KB}} = (K-1) \cdot \chi^2_{\mathbf{KB}} \\ &= K(K-1) \cdot N \cdot B^2 \cdot \text{Evenness}_{\mathbf{R}} \end{aligned} \quad (57)$$

This proves the Informedness form of the generalized $(K-1)^2$ degree of freedom $\chi^2$ statistic (42), and defines Evenness$_{\mathbf{R}}$ as the Arithmetic Mean of the individual dichotomous Evenness$_{\mathbf{R}}(l)$ terms (assuming B is constant). The Markedness form of the statistic (43) follows by analogous (Dual) argument, and the Correlation form (44) is simply the Geometric Mean of these two forms. Note however that this proof assumes that B is constant across all labels, and that assuming the determinant `det` is constant leads to a derivative of (20-21) involving a Harmonic Mean of Evenness as discussed in the next section.

The simplified $(K-1)$ degree of freedom $\chi^2_{\mathbf{K}}$ statistics were motivated as weighted averages of the dichotomous statistics, but can also be seen to approximate the $\chi^2_{\mathbf{X}}$ statistics given the observation that for a rejection threshold on the null hypothesis $H_0$, `alpha`< 0.05, the $\chi^2$ cumulative isodensity lines are locally linear in `r` (Fig. 4). Testing differences within a `beta` threshold as discussed above, is appropriate using the $\chi^2_{\mathbf{K}}$ series of statistics since they are postulated to have $(K-1)$ degrees of freedom. Alternately they may be tested according to the $\chi^2_{\mathbf{X}}$ series of statistics given they are postulated to differ in $(K-1)^2$ degrees of freedom, namely the noise, artefact and error terms that make the cells different between the two hypotheses (viz. that contribute to decorrelation). In practice, when used to test two systems or models other than the null, the models should be in a sufficiently linear part of the isodensity contour to be insensitive to the choice of statistic and the assumptions about degrees of freedom. When tested against the null model, a relatively constant error term can be expected to be introduced by using the lower degree of freedom model. The error introduced by the Cramer's V (K-1 degree of freedom) approximation to significance from $G^2$ or $\chi^2$ can be viewed in two ways. If we start with a $G^2$ or $\chi^2$ estimate as intended by Cramer we can test the accuracy of the estimate versus the true correlation, markedness and informedness as illustrated in Fig. 5. Note that we can see here that Cramer's V underestimates association for high levels of informedness, whilst it is reasonably accurate for lower levels. If we use (53) to (55) to estimate significance from the empirical association measures, we will thus underestimate significance under conditions of high association – viz. the test is more conservative as the effect size increases.

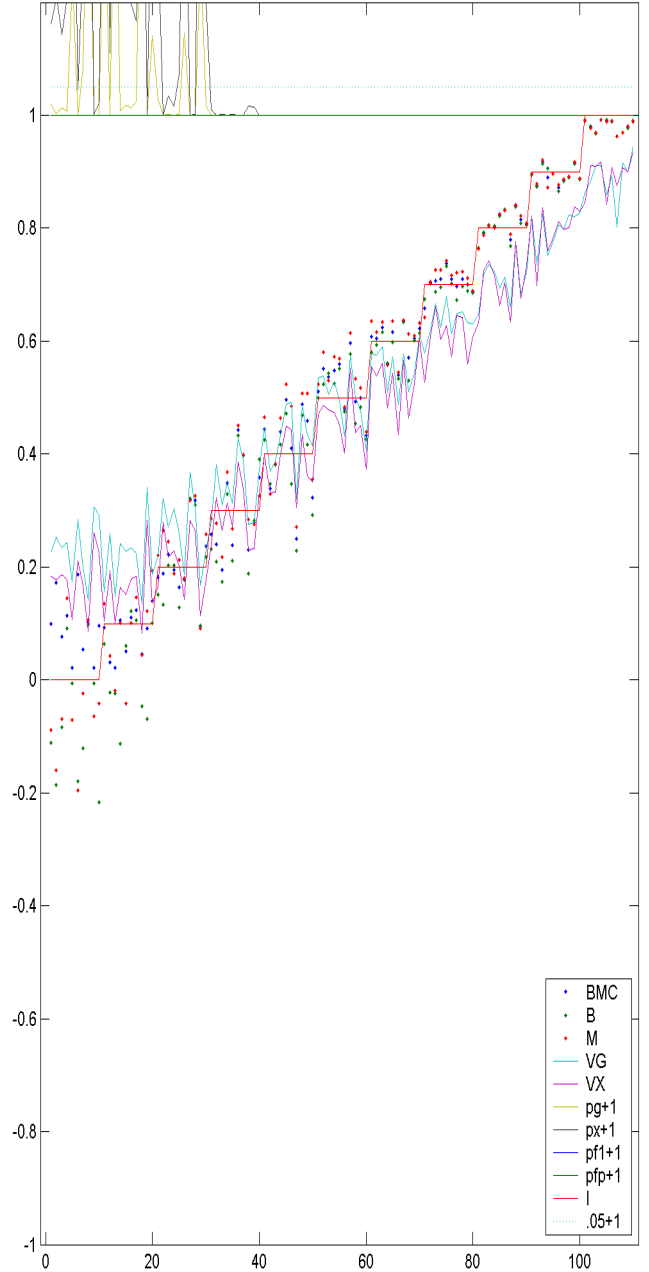

**Figure 5. Illustration of significance and Cramer's V.** 110 Monte Carlo simulations with 11 stepped expected Informedness (red) levels with Bookmaker-estimated Informedness (red dots), Markedness (green dot) and Correlation (blue dot), with significance (p+1) calculated using $G^2$, $X^2$, and Fisher estimates, and (skewed) Cramer's V Correlation estimates calculated from both $G^2$ and $X^2$. Here K=4, N=128, X=1.96, $\alpha=\beta=0.05$.

**Generalization of Evenness**

The proof that the product of dichotomous Evenness factors is the appropriate generalization in relation to the multiclass definition of Bookmaker Informedness and Markedness does not imply that it is an appropriate generalization of the dichotomous usage of Evenness in relation to Significance, and we have seen that the Arithmetic rather the Geometric Mean emerged in the above sketch proof. Whilst in general one would assume that Arithmetic and Harmonic Means approximate the Geometric Mean, we argue that the latter is the more appropriate basis, and indeed one may note that it not only approximates the Geometric Mean of the other two means, but is much more stable as the Arithmetic and Harmonic means can diverge radically from it in very uneven situations, and increasingly with higher dimensionality. On the other hand, the Arithmetic Mean is insensitive to evenness and is thus appropriate as a baseline in determining evenness. Thus the ratios between the means, as well as between the Geometric Mean and the geometric mean of the Arithmetic and Harmonic means, give rise to good measures of evenness.

On geometric grounds we introduced the Determinant of Correlation, `det`, generalizing `dp`, and representing the volume of possible deviations from chance covered by the target system and its perversions, showing its normalization to and Informedness-like statistic is Evenness$_{P+}$ the product of the Prevalences (and is exactly Informedness for `K=2`). This gives rise to an alternative dichotomous formulation for the aggregate false positive error for an individual case in terms of the `K-1` negative cases, using a ratio or submatrix determinant to submatrix product of Prevalences. This can be extended to all `K` cases while reflecting `K-1` degrees of freedom, by extending to the full contingency matrix determinant, `det`, and the full product of Prevalences, as our definition of another form of Evenness, Evenness$_{R\#}$

being the Harmonic Mean of the dichotomous Evenness terms for constant determinant:

$\chi^2_{KB}$ = `KN·det`$^{2/K}$ / Evenness$_{R\#}$ (58)
$\chi^2_{KM}$ = `KN·det`$^{2/K}$ / Evenness$_{P\#}$ (59)
$\chi^2_{KBM}$ = `KN·det`$^{2/K}$ / Evenness$_{G\#}$ (60)

Recall that the + form of Evenness is exemplified by

Evenness$_{R+}$ = $[\Pi_l\text{Prev}(l)]^{2/K}$ =PrevG (61)

and that the relationship between the three forms of Evenness is of the form

Evenness$_{R-}$ = Evenness$_{R+}$ / Evenness$_{R\#}$ (62)

where the + form is defined as the squared Geometric Mean (44-46), again suggesting that the – form is best approximated as an Arithmetic Mean (47-49). The above division by the Harmonic Mean is reminiscent of the Williams' correction which divides the $G^2$ values by an Evenness-like term `q=1+(a`$^2$`-1)/6Nr` where `a` is the number of categories for a goodness-of-fit test, `K` [18-20] or more generally, `K/PrevH` [17] which has maximum `K` when Prevalence is even, and `r=K-1` degrees of freedom, but for the more relevant usage as an independence test on a complete contingency table with `r=(K-1)`$^2$ degrees of freedom it is given by `a`$^2$`-1=(K/PrevH-1)·(K/BiasH-1)` where PrevH and BiasH are the Harmonic Means across the `K` classes or labels respectively [17-23].

In practice, any reasonable excursion from Evenness will be reflected adequately by any of the means discussed, however it is important to recognize that the + form is actually a squared Geometric Mean and is the product of the other two forms as shown in (62). An uneven bias or Prevalence will reduce all the corresponding Evenness forms, and compensate against reduced measures of association and significance due to lowered determinants.

Whereas broad assumptions and gross accuracy within an order of magnitude may be acceptable for calculating significance tests and p-values [23], it is clearly not appropriate for estimate the strength of associations. Thus the basic idea of Cramer's V is flawed given the rough assumptions and substantial errors associated with significance tests. It is thus better to start with a good measure of association, and use analogous formulae to estimate significance or confidence.

**Generalization of Confidence**

The discussion of confidence generalizes directly to the general case, with the approximation using Bookmaker Informedness[1], or analogously

---

[1] Informedness may be dichotomous and relates in this form to DeltaP, WRacc and the Gini Coefficient as discussed below. Bookmaker Informedness refers to the polychotomous generalization based on the Bookmaker analogy and algorithm [4].

Markedness, applying directly (the Informedness form is again a Prevalence weighted sum, in this case of a sum of squared versus absolute errors), viz.

$$CI_{B2} = X \cdot [1-|B|] / \sqrt{[2 E \cdot (N-1)]} \qquad (63)$$
$$CI_{M2} = X \cdot [1-|B|] / \sqrt{[2 E \cdot (N-1)]} \qquad (64)$$
$$CI_{C2} = X \cdot [1-|B|] / \sqrt{[2 E \cdot (N-1)]} \qquad (65)$$

In Equations 63-65 Confidence Intervals derived from the sse estimates of §2.8 are subscripted to show those appropriate to the different measures of association (Bookmaker Informedness, B; Markedness, M, and their geometric mean as a symmetric measure of Correlation, C). Those shown relate to `beta` (the empirical hypothesis based on the calculated B, giving rise to a test of power), but are also appropriate both for significance testing the null hypothesis (B=0) and provide tight (0-width) bounds on the full correlation (B=1) hypothesis as appropriate to its signification of an absence of random variation and hence 100% power (and extending this to include measurement error, discretization error, etc.)

The numeric subscript is 2 as notwithstanding the different assumptions behind the calculation of the confidence intervals (0 for the null hypothesis corresponding to `alpha`=0.05, 1 for the alternate hypothesis corresponding to `beta`=0.05 based on the weighted arithmetic model, and 2 for the full correlation hypothesis corresponding to `gamma`=0.05 – for practical purposes it is reasonable to use |1−B| to define the basic confidence interval for $CI_{B0}$, $CI_{B1}$ and $CI_{B2}$, given variation is due solely to unknown factors other than measurement and discretization error. Note that all error will lead to empirical estimates B<1.

If the empirical ($CI_{B1}$) confidence intervals include B=1, the broad confidence intervals ($CI_{B2}$) around a theoretical expectation of B=1 would also include the empirical contingency – it is a matter of judgement based on an understanding of contributing error whether the hypothesis B=1 is supported given non-zero error. In general B=1 should be achieved empirically for a true correlation unless there are measurement or labelling errors that are excluded from the informedness model, since B<1 is always significantly different from B=1 by definition (1−B=0 unaccounted variance due to guessing).

None of the traditional confidence or significance measures fully account for discretization error (N<8K) or for the distribution of margins, which are ignored by traditional approaches. To deal with discretization error we can adopt an sse estimate that is either constant independent of B, such as the unweighted arithmetic mean, or a non-trivial function that is non-zero at both B=0 and B=1, such as the weighted arithmetic mean:

$$CI_{B1} = X \cdot [1-2|B|+2B^2] / \sqrt{[2 E \cdot (N-1)]} \qquad (66)$$
$$CI_{M1} = X \cdot [1-2|B|+2B^2] / \sqrt{[2 E \cdot (N-1)]} \qquad (67)$$
$$CI_{C1} = X \cdot [1-2|B|+2B^2] / \sqrt{[2 E \cdot (N-1)]} \qquad (68)$$

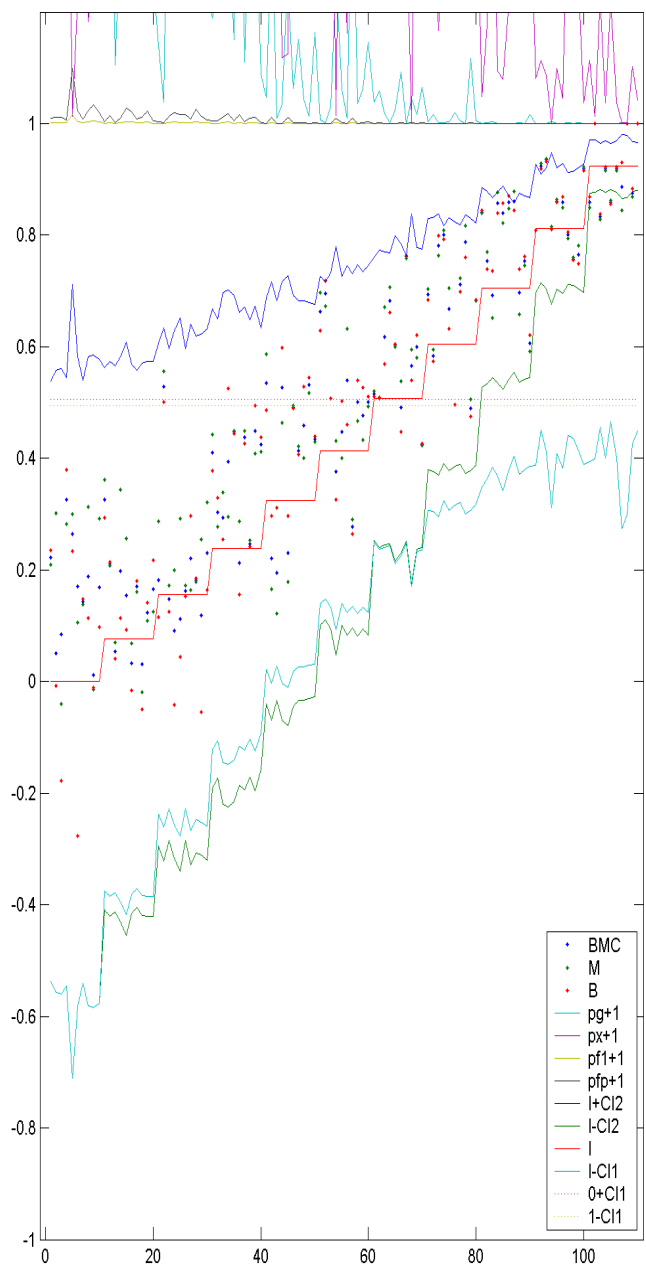

**Figure 6. Illustration of significance and confidence.**
110 Monte Carlo simulations with 11 stepped expected Informedness levels (red line) with Bookmaker-estimated Informedness (red dots), Markedness (green dot) and Correlation (blue dot), with significance (p+1) calculated using $G^2$, $X^2$, and Fisher estimates, and confidence bands shown for both the theoretical Informedness and the B=0 and B=1 levels (parallel almost meeting at B=0.5). The lower theoretical band is calculated twice, using both $CI_{B1}$ and $CI_{B2}$. Here K=4, N=16, X=1.96, α=β=0.05.

Substituting B=0 and B=1 into this gives equivalent CIs for the null and full hypothesis. In fact, it is sufficient to use the B=0 and 1 confidence intervals based on this variant since for X=2 they overlap at N<16. We illustrate such a marginal significance case in Fig. 6, where the large difference between the significance estimates is clear with Fisher showing marginal significance or better almost everywhere, $G^2$ for B>~0.6, $\chi^2$ for B>~0.8. >~95% of Bookmaker estimates are within the confidence bands as required (with 100% bounded by the more conservative lower band), however our B=0 and B=1 confidence intervals almost meet showing that we cannot distinguish intermediate B values other than B=0.5 which is marginal. Viz. we can say that this data seems to be random (B<0.5) or informed (B>0.5), but cannot be specific about the level of informedness for this small N (except for B=0.5±0.25).

If there is a mismatch of the marginal weights between the respective prevalences and biases, this is taken to contravene our assumption that Bookmaker statistics are calculated for the optimal assignment of class labels. Thus we assume that any mismatch is one of evenness only, and thus we set the Evenness factor E=PrevG*BiasG*$K^2$. Note that the difference between Informedness and Markedness also relates to Evenness, but Markedness values are likely to lie outside bounds attached to Informedness with probability greater than the specified `beta`. Our model can thus take into account distribution of margins provided the optimal allocation of predictions to categories (labelling) is assigned.

The multiplier X shown is set from the appropriate (inverse cumulative) Normal or Poisson distribution, and under the two-tailed form of the hypothesis, X=1.96 gives `alpha`, `beta` and `gamma` of 0.05. A multiplier of X=1.65 is appropriate for a one-tailed hypotheses at 0.05 level. Significance of difference from another model is satisfied to the specified level if the specified model (including null or full) does not lie in the confidence interval of the alternate model. Power is adequate to the specified level if the alternate model does not lie in the confidence interval of the specified model. Figure 7 further illustrates the effectiveness of the 95% empirical and theoretical confidence bounds in relation to the significance achievable at N=128 (K=5).

**EXPLORATION AND FUTURE WORK**

Powers Bookmaker Informedness has been used extensively by proponent and his students over the last 10 years, in particular in the PhD Theses and other publications relating to AudioVisual Speech Recognition [25-26] and EEG/Brain Computer Interface [27-28], plus Matlab scripts that are available for calculating both the standard and Bookmaker statistics[2] (these were modified by the present author to produce the results presented in this paper). The connection with DeltaP was noted in the course of collaborative research in Psycholinguistics, and provides an important Psychological justification or confirmation of the measure where biological plausibility is desired. We have referred extensively to the equivalence of Bookmaker Informedness to ROC AUC, as used standardly in Medicine, although AUC has the apparent form of an undemeaned probability based on a parameterized classifier or a series of classifiers, and B is a demeaned renormalized kappa-like form based on a single fully specified classifier.

---

[2] http://www.mathworks.com/matlabcentral/fileexchange/5648-informedness-of-a-contingency-matrix

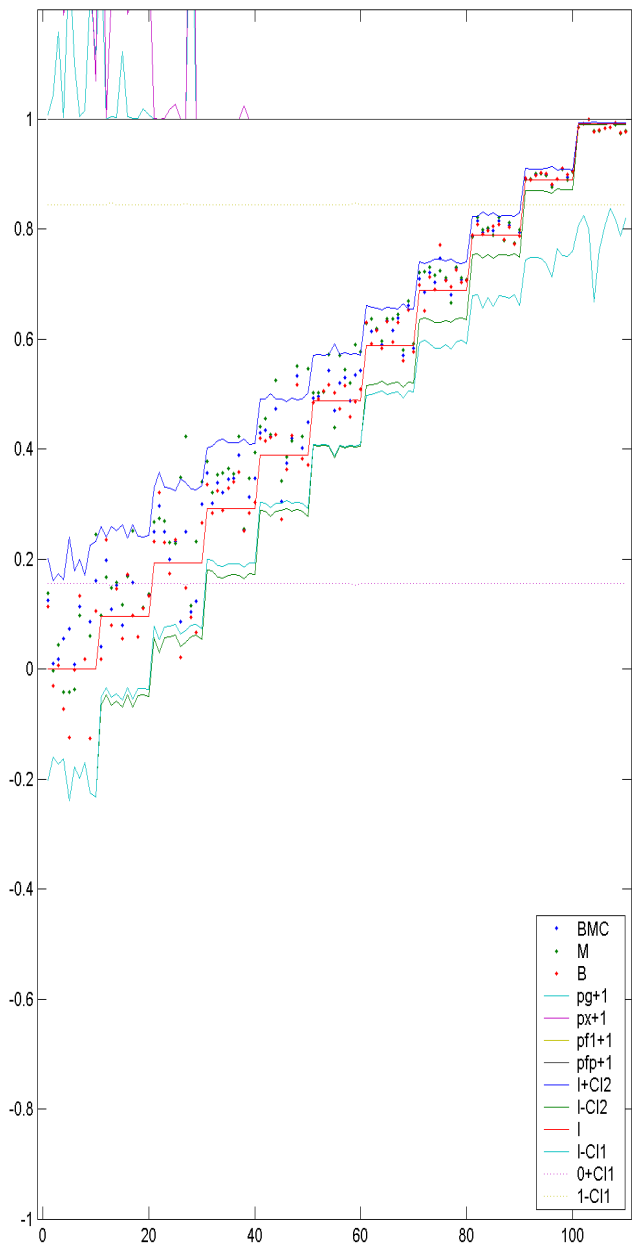

**Figure 7. Illustration of significance and confidence.**
110 Monte Carlo simulations with 11 stepped expected Informedness levels (red line) with Bookmaker-estimated Informedness (red dots), Markedness (green dot) and Correlation (blue dot), with significance (p+1) calculated using $G^2$, $X^2$, and Fisher estimates, and confidence bands shown for both the theoretical Informedness and the B=0 and B=1 levels (parallel almost meeting at B=0.5). The lower theoretical band is calculated twice, using both $CI_{B1}$ and $CI_{B2}$. Here K=5, N=128, X=1.96, α=β=0.05.

The Informedness measure has thus proven its worth across a wide range of disciplines, at least in its dichotomous form. A particular feature of the major studies that used Informedness, is that they covered different numbers of classes (exercising the multi-class form of Bookmaker as implemented in Matlab), as well as a number of different noise and artefact conditions. Both of these aspects of their work meant that the traditional measures and derivatives of Recall, Precision and Accuracy were useless for comparing the different runs and the different conditions, whilst Bookmaker gave clear unambiguous, easily interpretable results which were contrasted with the traditional measures in these studies.

The new $\chi^2_{KB}$, $\chi^2_{KM}$ and $\chi^2_{KBM}$, $\chi^2_{XB}$, $\chi^2_{XM}$ and $\chi^2_{XBM}$ correlation statistics were developed heuristically with approximative sketch proofs/arguments, and have only been investigated to date in toy contrived situations and the Monte Carlo simulations in Figs 2, 3, 5, 6 and 7. In particular, whilst they work well in the dichotomous state, where they demonstrate a clear advantage over traditional $\chi^2$ approaches, there has as yet been no application to our multi-class experiments and no major body of work comparing new and conventional approaches to significance. Just as Bookmaker (or DeltaP') is the normative measure of accuracy for a system against a Gold Standard, so is $\chi^2_{XB}$ the proposed $\chi^2$ significance statistic for this most common situation in the absence of a more specific model (noting that $\sum x = \sum x^2$ for dichotomous data in {0,1}). For the cross-rater or cross-system comparison, where neither is normative, the BMG Correlation is the appropriate measure, and correspondingly we propose that $\chi^2_{KBM}$ is the appropriate $\chi^2$ significance statistic. To explore these thoroughly is a matter for future research. However, in practice we tend to recommend the use of Confidence Intervals as illustrated in Figs 4 and 5, since these give a direct indication of power versus the confidence interval on the null hypothesis, as well as power when used with confidence intervals on an alternate hypothesis.

Furthermore, when used on the empirical mean (correlation, markedness or informedness), the overlap of the interval with another system, and vice-versa, give direct indication of both significance and power of the difference between them. If a system occurs in another confidence interval it is not significantly different from that system or hypothesis, and if it is it is significantly different. If its own confidence interval also avoids overlapping the alternate mean this mutual significance is actually a reflection of statistical power at a complementary level. However, as with significance tests, it is important to avoid reading to avoid too much into non-overlap of interval and mean (not of intervals) as the actual probabilities of the hypotheses depends also on unknown priors.

Thus whilst our understanding of Informedness and Markedness as performance measure is now quite mature, particularly in view of the clear relationships with existing measures exposed in this article, we do not regard current practice in relation to significance and confidence, or indeed our present discussion, as having the same level of maturity and a better understanding of the significance and confidence measures remains a matter for further work, including in particular, research into the multi-class application of the technique, and exploration of the asymmetry in degrees of freedom appropriate to `alpha` and `beta`, which does not seem to have been explored hitherto. Nonetheless, based on pilot experiments, the dichotomous $\chi^2_{KB}$ family of statistics seems to be more reliable than the traditional $\chi^2$ and $G^2$ statistics, and the confidence intervals seem to be more reliable than both. It is also important to recall that the marginal assumptions underlying the both the $\chi^2$ and $G^2$ statistics and the Fisher exact test are not actually valid for contingencies based on a parameterized or learned system (as opposed to naturally occurring pre- and post-conditions) as the different tradeoffs and algorithms will reflect different margins (biases).

It also remains to explore the relationship between Informedness, Markedness, Evenness and the Determinant of Contingency in the general multiclass case. In particular, the determinant generalizes to multiple dimensions to give a volume of space that represents the coverage of parameterizations that are more random than contingency matrix and its perverted forms (that is permutations of the classes or labels that make it suboptimal or subchance). Maximizing the determinant is necessary to maximize Informedness and Markedness and hence Correlation, and the normalization of the determinant to give those measures as defined by (42-43) defines respective multiclass Evenness measures satisfying a generalization of (20-21). This alternate definition needs to be characterized, and is the exact form that should be used in equations 30 to 46. The relationship to the discussed mean-based definitions remains to be explored, and they must at present be regarded as approximative. However, it is possible (and arguably desirable) to instead of using Geometric Means as outlined above, to calculate Evenness as defined by the combination of (20-22,42-43). It may be there is an simplified identity or a simple relationship with the Geometric Mean definition, but such simplifications have yet to be investigated.

**MONTE CARLO SIMULATION**

Whilst the Bookmaker measures are exact estimates of various probabilities, as expected values, they are means of distributions influenced not only by the underlying decision probability but the marginal and joint distributions of the contingent variables. In developing these estimates a minimum of assumptions have been made, including avoiding the assumption that the margins are predetermined or that bias tracks prevalence, and thus it is arguable that there is no attractor at the expected values produced as the independent product of marginal probabilities. For the purposes of Monte Carlo simulation, these have been implemented in Matlab 6R12 using a variety of distributions across the full contingency table, the uniform variant modelling events hitting any cell with equal probability in a discrete distribution with $K^2-1$ degrees of freedom (given N is fixed). In practice, (pseudo-)random number will not automatically set $K^2$ random numbers so that they add exactly to N, and setting $K^2-1$ cells and allowing the final cell to be determined would give it o(K) times the standard deviation of the other cells. Thus another approach is to approximately specify N and either leave the number of elements as it comes, or randomly increment or decrement cells to bring it back to N, or ignore integer discreteness constraints and renormalize by multiplication. This raises the question of what other constraints we want to maintain, e.g. that cells are integral and non-negative, and that margins are integral and strictly positive.

An alternate approach is to separately determine the prediction bias and real prevalence margins, using a uniform distribution, and then using conventional distributions around the expected value of each cell. If we believe the appropriate distribution is normal, or the central limit applies, as is conventionally assumed in the theory of $\chi^2$ significance as well as the theory of confidence intervals, then a normal distribution can be used. However, if as in the previous model we envisage events that are allocated to cells with some probability, then a binomial distribution is appropriate, noting that this is a discrete distribution and that for reasonably large N it approaches the normal distribution, and indeed the sum of independent events meets the definition of the normal distribution except that discretization will cause deviation. In general, it is possible that the expected marginal distribution is not met, or in particular that the assumption that no marginal probability is 0 is not reflected in the empirical distributions.

Monte Carlo simulations have been performed in Matlab using all the variants discussed above. Violating the strictly positive margin assumption causes NaNs for many statistics, and for this reason this in enforced by setting 1s at the intersection of paired zero-margin rows and columns, or arbitrarily for unpaired rows or columns. Another way of avoiding these NaN problems is to relax the integral/discreteness assumptions. Uniform margin-free distribution, discrete or real-valued, produces a broader error distribution than the margin-constrained distributions. It is also possible to use so-called copula techniques to reshape uniformly distributed random numbers to another distribution. In addition Matlab's directly calculated `binornd` function has been used to simulate the binomial distribution, as well as the absolute value of the normal distribution shifted by (plus) the binomial standard deviation. No noticeable difference has been observed due to relaxing the integral/discreteness assumptions except for disappearance of the obvious banding and more prevalent extremes at low N, outside the recommended minimum expected count of 5 per cell for significance and confidence estimates to be valid. On the other hand, we note that the built in `binornd` produced unexpectedly low means and always severely underproduced before correction[3]. This leads to a higher discretization effect and less randomness, and hence overestimation of associations. The direct calculation over N events means it takes o(N) times longer to compute and is impractical for N in the range where the statistics are meaningful. The `binoinv` and related functions ultimately use `gammaln` to calculate values and thus the copula technique is of reasonable order, its results being comparable with those of absolute normal.

Figures 2, 3, 5, 6 and 7 have thus all been based on pre-marginalized simulations in Matlab, with discretized absolute normal distributions using post-processing as discussed above to ensure maintenance of all constraints, for K=2 to 102 with expected value of N/K = $2^1$ to $2^9$ and expected B of 0/10 to 10/10, noting that the forced constraint process introduces additional randomness and that the relative amount of correction required is expected to decrease with K.

**CONCLUSIONS**

The system of relationships we have discovered is amazingly elegant. From a contingency matrix in count or reduced form (as probabilities), we can construct both dichotomous and mutually exclusive multiclass statistics that correspond to debiased versions of Recall and Precision (28,29). These may be related to the Area under the Curve and distance from (1,1) in the Recall-based ROC analysis, and its dual Precision-based method, for a single fully-specified classifier (viz. after fixing threshold and/or other parameters). There are further insightful relationships with Matthews Correlation, with the determinant of either form of the matrix (`DTP` or `dtp`), and the Area of the Triangle defined by the ROC point and the chance line, or equivalently the Area of the Parallelogram or Trapezoid defined by its perverted forms.

---

[3] The author has since found and corrected this Matlab initialization bug.

Also useful is the direct relationship of the three Bookmaker goodness measures (Informedness, Markedness and Matthews Correlation) with both standard (biased) single variable significance tests as well as the clean generalization to unbiased significance tests in both dependent (low degree of freedom) and independent (high degree of freedom) forms along with simple formulations for estimating confidence intervals. More useful still is the simple extension to confidence intervals which have the advantage that we can compare against models other than the null hypothesis corresponding to B=0. In particular we also introduce the full hypothesis corresponding full informedness at B=1 mediated by measurement or labelling errors, and can thus distinguish when it is appropriate to recognize a specific value of partial informedness, 0<B<1 (which will eventually be the case for any association that isn't completely random, for large enough N).

It is also of major importance that the measures are easily generalized to multiclass contingency tables. The multiclass form of the Informedness measure has been used extensively as the primary goodness measure in a number of papers and PhD theses in different areas of Artificial Intelligence and Cognitive Science (Matlab scripts are available through Matlab Central[2]), and in Psychology the pair of dichotomous measures, under the names DeltaP and DeltaP' have been explored extensively and shown empirically to be normative measures of human associative performance [9].

Most encouraging of all is how easy the techniques are to teach and are conceptuatlly and programmatically simpler than RoC – they are taught routinely to Honours students and used routinely by all students in our lab, and they directly give probabilities regarding the effectiveness of the system. The dichotomous forms are trivial: Informedness is simply Recall plus Inverse Recall minus 1(or equivalently Sensitivity + Specificity – 1), and Markedness is Precision plus Inverse Precision minus 1. The observation that their Geometric Mean is Matthews Correlation also answers questions about how best to reduce these pairs to the form of a single probability, and they can also be expressed as a demeaned average. Evenness is the square of the Geometric Mean of Prevalence and Inverse Prevalence and/or Bias and Inverse Bias. $\chi^2$ testing is just multiplication by a constant, and conservative confidence intervals are then a matter of taking a squareroot.

There is also an intuitive relationship between the unbiased measures and their significance and confidence, and we have sought to outline a rough rationale for this, but this remains somewhat short of formal proof of optimal formulae defining close bounds on significance and confidence.

**FUTURE WORK**

The major area of future work is the development and exploration of learning algorithms that optimize a chance-correct measure. This is a special case of the optimization of a given cost measure, where the cost is given by Bookmaker odds, that is the prevalence of wins for each class. One of the most important insights from the Bookmaker analogy is the dichotomous Informedness estimates need to be averaged weighted by the bias, that is the number of times a label is predicted and bet on – it's contribution to your overall payoff is a function of how often you bet on it, not how often it happens.

Current work is exploring chance-correct evaluation in modifications to neural network, machine learning and boosting optimization. If you are optimizing the wrong thing, you will not do as well as if you optimize Informedness, and optimizing Informedness will also optimize uncorrected accuracy and error measures.

Recent work has demonstrated that using optimizing appropriate measures can make a huge difference to the final accuracies achieved. This has been particularly clear in a boosting context [29-30].